\documentclass[10pt,twocolumn,letterpaper]{article}

\usepackage{iccv}
\usepackage{times}
\usepackage{epsfig}
\usepackage{graphicx}
\usepackage{amsmath}
\usepackage{amssymb}
\usepackage{booktabs}
\usepackage{mathrsfs}
\usepackage{bm}
\usepackage{algorithm, algorithmic}
\usepackage{mathdots}
\usepackage{multirow}
\usepackage[american]{babel}
\usepackage{microtype} 
\usepackage{array}
\usepackage{setspace}
\usepackage{threeparttable}
\usepackage{graphicx}
\usepackage{diagbox}
\usepackage{wrapfig}
\usepackage{float}
\usepackage{bbm}
\usepackage{url}

\usepackage[breaklinks=true,bookmarks=false]{hyperref}

\iccvfinalcopy 

\def\iccvPaperID{6021} 

\ificcvfinal\fi

\begin{document}

\title{Controllable Guide-Space for Generalizable Face Forgery Detection}


\author{
Ying Guo \thanks{Equal Contribution. $^\dag$Corresponding author.} \ ,  Cheng Zhen \footnotemark[1] \ ,  Pengfei Yan \textsuperscript{\dag} \\
Vision AI Department, Meituan\\
{\tt\small  \{guoying16, zhencheng02, yanpengfei03\}@meituan.com }
}
\maketitle
\pagestyle{plain}

\begin{abstract}
Recent studies on face forgery detection have shown satisfactory performance for methods involved in training datasets, but are not ideal enough for unknown domains.
This motivates many works to improve the generalization, but forgery-irrelevant information, such as image background and identity, still exists in different domain features and causes unexpected clustering, limiting the generalization.
In this paper, we propose a controllable guide-space (GS) method to enhance the discrimination of different forgery domains, so as to increase the forgery relevance of features and thereby improve the generalization.
The well-designed guide-space can simultaneously achieve both the proper separation of forgery domains and the large distance between real-forgery domains in an explicit and controllable manner. 
Moreover, for better discrimination, we use a decoupling module to weaken the interference of forgery-irrelevant correlations between domains.
Furthermore, we make adjustments to the decision boundary manifold according to the clustering degree of the same domain features within the neighborhood.
Extensive experiments in multiple in-domain and cross-domain settings confirm that our method can achieve state-of-the-art generalization.
\end{abstract}

\section{Introduction}
\vspace{-0.1cm}
Face forgery technology \cite{deepfake, fs, infoswap} has made vigorous development in recent years.
However, these realistic forgery faces are sometimes abused to maliciously disguise identities, especially celebrities and politicians, causing serious social problems.
Therefore, how to reduce this risk has attracted widespread attention from researchers.

Convolutional neural networks (CNNs) have shown excellent performance in face forgery detection \cite{attention1, twostream, xception, sola}. According to forgery or not, this task is often formalized as a binary classification problem, and some suitable classification networks \cite{xception, capsule} are introduced to this task.
Although they perform well in the training domain, the learned features may be method-specific for the forgery methods within the training set \cite{recce}, and cannot show satisfactory generalization in unknown forgery methods.


\begin{figure}[t]
\centering\includegraphics[width=0.42\textwidth]{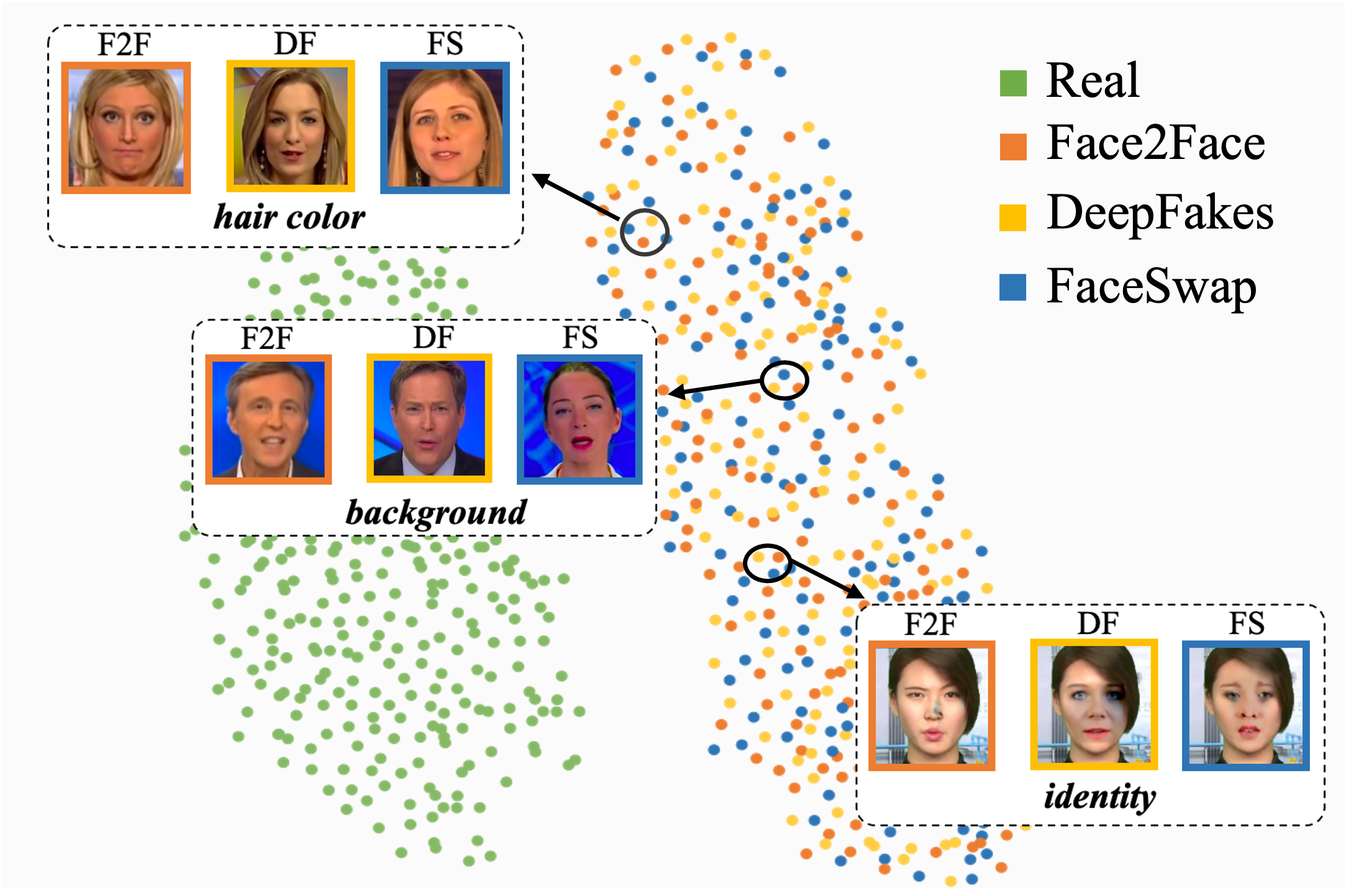}
\caption{Limitations of existing methods: features from different domains are still clustered by forgery-irrelevant similarities (image background, hair color, identity, etc.), proving that features still contain forgery-irrelevant information, limiting the generalization.}
\label{fig:intro}
\vspace{-0.27cm}
\end{figure}

Forgery data generated from various methods corresponds to different forgery domains.
To improve the generalization in unknown domains, 
some works \cite{facexray,selfblended,advtrain_cvpr22} study the common artifacts of various forgeries and use data augmentation to synthesize more training data.
Several other works are devoted to mining better discriminative features, and attention mechanisms \cite{mat}, local relation \cite{localrelation}, and frequency information \cite{scl, f3net} are also introduced to capture better forgery traces.
SRM \cite{srm} suppresses the acquisition of color and texture via high-frequency noise, thereby solving the overfitting to the training data.
In addition, RECCE \cite{recce} copes with the complexity of various forgery domains by learning compact real representations based on the reconstruction.
Compared with previous methods, the exploration of common characteristics and discriminative information allows the model to learn relatively more generalized features.

Nevertheless, during training, the model tries to increase the discrimination between real and fake features, but treats different forgery domains (i.e. different forgery methods) as the unified ``fake" category without distinction.
In common training sets \cite{ff++,celebdf}, there are some similarities between massive data in forgery-irrelevant information, such as hair color, image background, and identity. 
Due to the uniform fake categorization, the goal of training is only to distinguish the fake from the real, without making further distinctions between forgery types.
As a result, as shown in Figure \ref{fig:intro}, some features will present a clustering phenomenon based on above forgery-irrelevant similarities (hair color, background, identity, etc.) rather than forgery domain characteristics that are more relevant to the forgery detection task. This demonstrates that the learned features inevitably still contain some forgery-irrelevant information \cite{shapes}.

\begin{figure}[t]
\centering\includegraphics[width=0.48\textwidth]{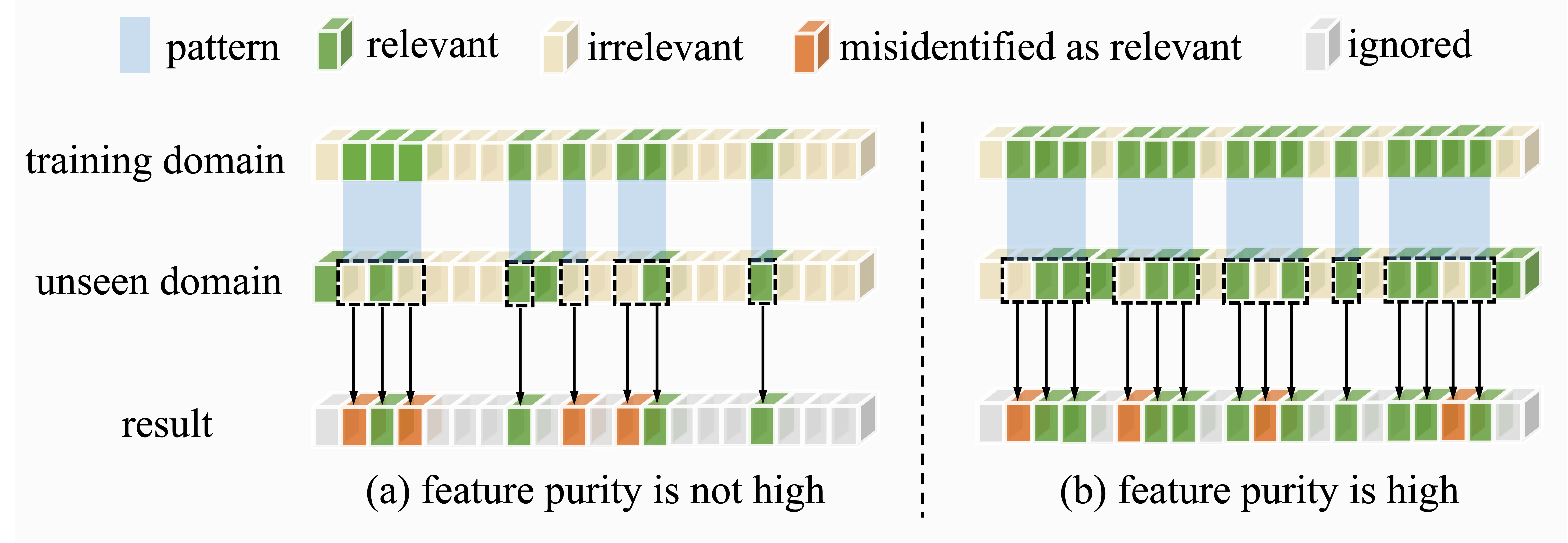}
\caption{Comparisons of the forgery-relevance for model focusing results at different feature purities.}
\label{fig:pure}
\vspace{-0.27cm}
\end{figure}

The mixing of irrelevant information in features (i.e. the feature purity is not high) may limit the generalization.
As shown in Figure \ref{fig:pure} (a), in the training domain, guided by the supervised information, the model will learn a pattern of which features are more relevant, and show good performance in the current domain.
However, in unseen domains, the feature distribution has a deviation. 
Following the original pattern, the features that the model focuses on will contain a high proportion of irrelevant information, which may cause the model to make decisions based on the similarity of such irrelevant information.
On the contrary, as shown in Figure \ref{fig:pure} (b), when the feature purity is high, the irrelevant information contained in the feature itself is less. Even if the distribution of the unseen domain is biased, the proportion of irrelevant information in the extracted features will be correspondingly less.
So we believe that higher feature purity will help the generalization, and we also give a theoretical proof in Appendix 1.

Based on above considerations, in this paper, instead of treating all forgery types as a unified category, we propose a novel guide-space (GS) based framework to increase a proper level of discrimination between different forgery domains.
In this way, by learning the differences between forgery domains and the consistency of the same domain, the model can further pay more attention to forgery traces.
And separating the features of different domains can reduce their correlation in irrelevant information.
The learned features are more forgery-relevant, thus helping the generalization.

Specifically, the increase in forgery domain discrimination needs to be controlled within a certain range, because a larger real-forgery distance should be preferably maintained at the same time.
In this way, in unseen domains, the forgery features will be located far away from real features with a higher probability.
Thus in our guide-space, we construct the guide embeddings of the real and different forgery domains, and make the features approach their respective guide embeddings to actively control the compactness of the real domain and the separation degree between different forgery domains.
Further, considering that the correlation between different domains in terms of forgery-irrelevant similarity will interfere with the domain distinction, we mine this potential correlation based on the clustering results of the self-supervised features of images, and decouple the irrelevant information accordingly.
In addition, we design a decision boundary manifold adjustment module (A-DBM) based on the degree of feature aggregation, to better realize the feature distribution defined by the guide-space.

In summary, this paper has the following contributions:
\begin{itemize}
\setlength{\itemsep}{0pt}
\vspace{-0.17cm}
\item We argue that a proper level of discrimination between different forgery domains is also important to improve the generalization, so as to capture more forgery-relevant information and to weaken the impact of forgery-irrelevant information.
\vspace{-0.1cm}
\item We construct a guide-space to achieve the controllable separation of both real-forgery domains and forgery-forgery domains, and further decouple the forgery-irrelevant correlation between different domains to reduce their interference on domain separation.
\vspace{-0.1cm}
\item We design an adjustment strategy for the decision boundary manifold to make the features of the same domain better clustered and compliant with the distribution of the guide-space.
\vspace{-0.1cm}
\item Extensive experiments in multiple cross-domain settings confirm that our method can realize the state-of-the-art generalization, and achieve the cross-domain AUC of 84.97\% and 81.65\% on CelebDF and DFDC.
\end{itemize}

\section{Related works}
Face forgery detection based on convolutional neural networks (CNNs) has been widely used \cite{twostream,mesonet,lips,geometry,twobranch,osn,fretal,multitask,uia-vit}.
Early works \cite{xception,capsule} apply suitable classification networks to forgery detection tasks, and achieve good performance on the forgery domains presented in the training set. However, the learned features may be more suitable for forgery methods presented in the training set \cite{recce}, and cannot show good generalization on unknown forgery domains.

\begin{figure*}[t]
\centering\includegraphics[width=0.9\textwidth]{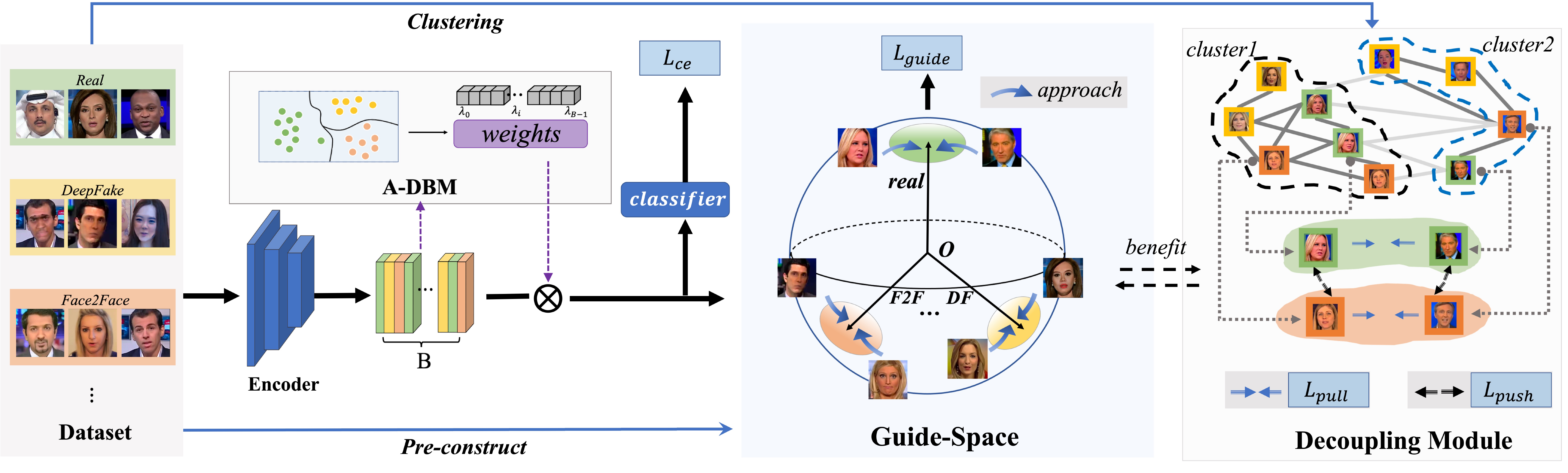}
\caption{Overview of our framework, including guide-space based controllable optimization, adjustment of decision boundary manifold (A-DBM), and irrelevant information decoupling module.}
\label{fig:frame}
\end{figure*}

Recently, more efforts have been made to improve this generalization.
Some methods attempt to learn common characteristics of different forgery domains.
For example, works in \cite{facexray,selfblended,rl_aug,advtrain_cvpr22} use data augmentation to simulate common artifacts (blending boundary, color inconsistency, etc.) of forgeries.
SPSL \cite{spsl} captures the phase spectrum changes caused by common up-sampling operations during the forgery process.
They exhibit improved generalization, but these common features often cover a limited variety of forgeries.
Several other studies are devoted to mining better discriminative features.
To better capture forgery clues, the attention mechanism \cite{mat,attention1,rfm}, amplification strategy \cite{sola}, or local relationships \cite{PCLI2G,localrelation} between regions are studied.
Besides low-level RGB features, the frequency information \cite{f3net,srm,frepgan,scl} is also introduced.
In addition, RECCE \cite{recce} based on reconstruction learning tries to learn compact representations of real data to cope with the complexity of forgery domains. LTW \cite{ltw} utilizes meta-learning to balance the performance across multiple domains.


Ideally, for a generalized model that extracts forgery-related information, the forgery features should be aggregated according to their respective domain types rather than forgery-irrelevant information. However, although the above methods mine better forgery traces, they treat different forgery domains as the uniform ``fake", which makes the features with forgery-irrelevant similarities but belonging to different domains still cluster together.



\section{Methodology}\label{sec:method}
\vspace{-0.15cm}
To improve the generalization, we propose a novel guide-space (GS) based framework, which consists of three main schemes, i.e., guide-space based controllable optimization, adjustment of decision boundary manifold (A-DBM), and decoupling module for irrelevant information, as illustrated in Figure \ref{fig:frame}.
We first pre-construct an ideal guide-space, making features closer to their guide embeddings of respective domains. To better aggregate features of the same domain, we adjust the decision boundary manifold by setting weights of samples within a batch. Further, to mitigate the interference of irrelevant correlations, we decouple these correlations with the aid of self-supervised feature clustering.
The guide-space and decoupling module can benefit from each other to make features achieve better forgery relevance.
The following subsections show details of three schemes.
\vspace{-0.25cm}
\subsection{Construction of guide-space}\label{sec:guide}
\vspace{-0.15cm}
Before training, we first construct a guide-space containing guide embeddings for the real domain and different types of forgery domains. 
In the subsequent training, the features of different domains are approached to their respective guide embeddings to achieve distinguishability from each other.

\begin{figure}[t]
\centering\includegraphics[width=0.35\textwidth]{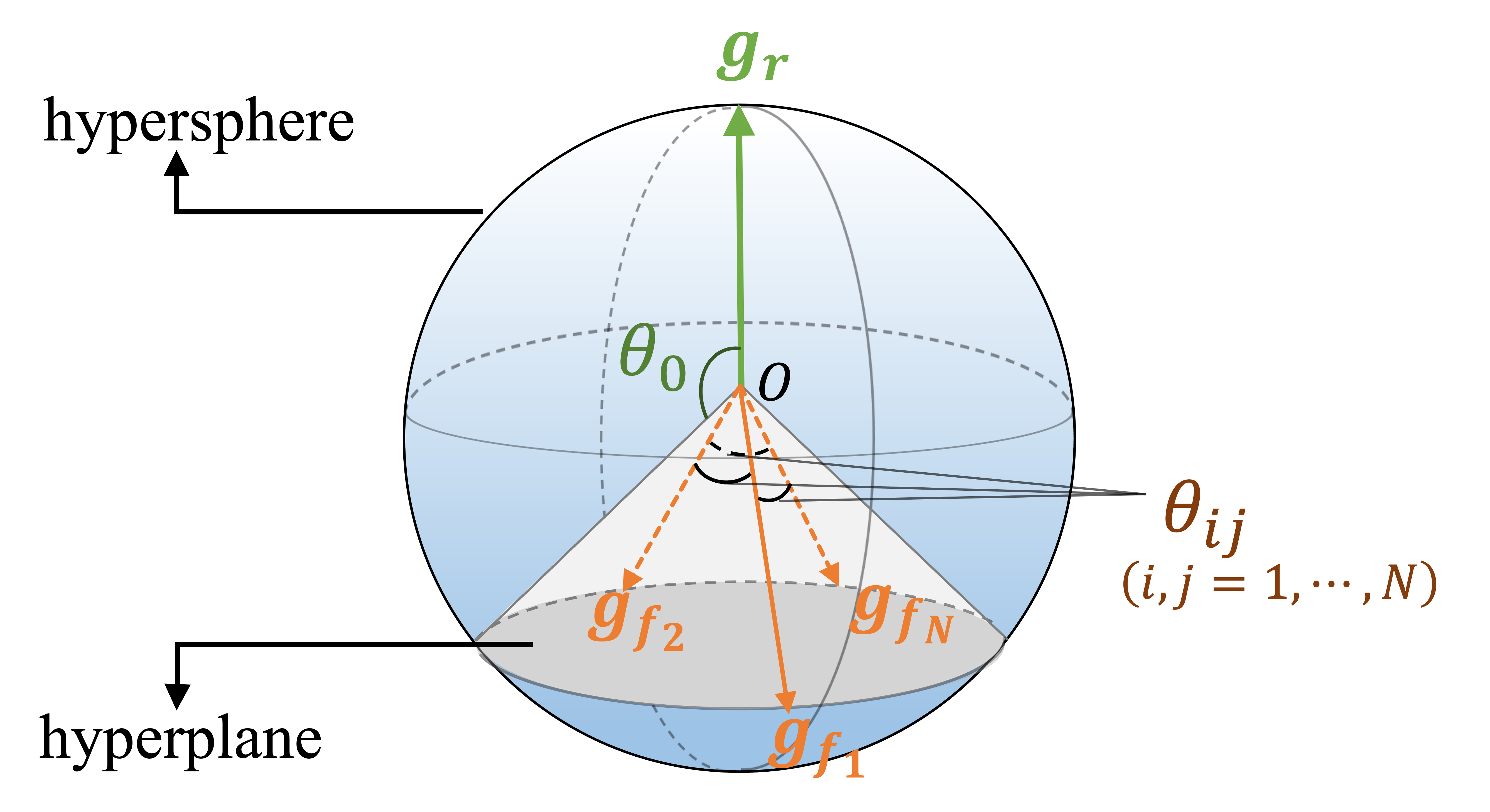}
\caption{Visualization of the guide-space. The real guide embedding $\bm{g_r}$ and all forgery guide embeddings $\bm{g_{_f}}$ are at a fixed angle $\theta_0$ ($\theta_0 > \theta_{ij}$), and the larger the angle $\theta_{ij}$ between $\bm{g_{_f}}$, the better.}
\label{fig:sphere}
\end{figure}

The construction of the guide-space requires the dimension $d$ of the face feature representation and the number of forgery categories $N$ in the training set. The features lie on a hypersphere of unit length $S=\left\{v \in R^d \mid\|v\|=1\right\}$. Let $\bm{g_r}$ and $\bm{g_{_{f}}}=\left\{\bm{g_{_{f_i}}} \mid i=1, \!\cdots \!, N\right\}$ represent the guide embeddings of the real and forgery domains to be solved, respectively. The visualization of $\bm{g_r}$ and $\bm{g_{_f}}$ in the guide-space is shown in Figure \ref{fig:sphere}.

On the one hand, we make $\bm{g_r}$ and all embeddings in $\bm{g_{_f}}$ present a fixed large angle $\theta_0$ (a hyperparameter) to separate real and fake features. $\theta_0$ can explicitly and actively control the separation degree between real-forgery domains, and a large $\theta_0$ ensures the compactness of real domain features:
\vspace{-0.17cm}
\begin{equation}\label{eq:theta0}
e^{\bm{g_r}^{T} \bm{g_{_{f_i}}}}=e^{cos(\theta_0)}(i=1, \cdots, N)
\vspace{-0.1cm}
\end{equation}

\vspace{-0.15cm}
On the other hand, the restriction of $\theta_0$ makes the embeddings in $\bm{g_{_f}}$ be located in a hyperplane of $d-1$ dimensions. Then all embeddings in $\bm{g_{_f}}$ should be as far as possible from each other, so as to increase the discrimination between forgery domains, and weaken the similarities in forgery-irrelevant information. This optimization is formulated as:
\vspace{-0.18cm}
\begin{equation}\label{eq:l_cf}
L\left(\left\{\bm{g_{_{f_i}}}\right\}_{i=1}^N\right)=\frac{1}{N} \sum_{i=1}^N \log \sum_{j=1}^N e^{\bm{g_{{}_{f_i}}}^T \bm{g_{_{f_j}}} / \tau}
\vspace{-0.2cm}
\end{equation}
where $\tau$ is a temperature parameter to control the scale of distribution \cite{dcl}.
We first obtain $\bm{g_r}$ by random initialization, and then take Eq. (\ref{eq:theta0}) as the constraint of Eq. (\ref{eq:l_cf}) and solve this constrained optimization problem according to the Lagrangian multiplier method \cite{lagrangian} to obtain $\bm{g_{_f}}$.
The solving process is detailed in Appendix 2.
To guarantee that the equation is solvable, $d \geq N$.

Let $\theta_{ij}(i, j=1, \cdots, N, i \neq j)$ represent the angle between $\bm{g_{_f}}$. 
Adjusting $\theta_0$ can affect $\theta_{ij}$ and thus adjust the degree of separation between all embeddings.
Note that $\theta_0$ is not as large as possible, because in the space with limited dimensions, the larger $\theta_0$, the smaller $\theta_{ij}$, and the separation $\theta_{ij}$ between forgery domains also needs to be maintained.
With this well-designed guide-space, we can achieve both the separation of forgery domains and the compactness of the real domain. More importantly, this process is explicit and controllable, rather than implicit and uncontrolled learning.
\vspace{-0.05cm}
\subsection{Controllable optimization based on guide-space}
\vspace{-0.09cm}
In the optimization, let $\left\{\left(x_i, y_i, t_i\right)\right\}_{i=1}^B$ denote a batch of face images, where $y_i$ is the ground-truth label that marks whether the image is fake or not, i.e., $y_i \in\{0,1\}$. $t_i$ refers to the domain label, i.e., $t_i \in[0,N]$, where for real faces, $t_i=0$, and for fake faces, $t_i$ represents the forgery category label to which it belongs.
The forgery detection model consists of a feature encoder $F(\cdot)$ followed by a binary classifier $h(\cdot)$.

Based on the pre-calculated guide embeddings in Sec. \ref{sec:guide}, we make the features of each domain close to their respective guide embeddings, so as to achieve the separation between real-forgery domains and between forgery-forgery domains.
Let $v_i$ denote the feature of image $x_i$ extracted by $F(\cdot)$, $G$ denote the set of all guide embeddings, $G=\left\{\bm{g_r}, \bm{g_{_{f_i}}} \mid i=1, \!\cdots \!, N\right\}$, and the loss function can be formulated as:
\vspace{-0.12cm}
\begin{equation}\label{eq:l_guide}
L_{guide }=-\sum_{i=1}^B \lambda_i \log \frac{e^{v_i^T g_i^* / \tau}}{\sum_{v_j \in V \cup G} e^{v_i^T v_j / \tau}}
\vspace{-0.1cm}
\end{equation}
where $\lambda_i$ is the weight of the current data $x_i$ in the loss calculation relative to the data within a batch, and in general, the loss is the average of each data, i.e., $\lambda_i=1 / B$. $g_i^*$ is the guide embedding corresponding to $x_i$. $g_i^*=\left\{\begin{array}{cl}g_r & \text { if } t_i=0 \\ g_{_{f_j}}\left(j=\Phi\left(t_i\right)\right) & \text { if } 1 \leq t_i \leq N\end{array}\right.$. $\Phi(\cdot)$ is the relation function between forgery domains and forgery guide embeddings.
At each iteration, we compute the average feature for each forgery domain. According to the distance between each average feature and the guide embedding, we use Hungarian algorithm \cite{hungarian} to perform nearest-neighbor matching, and denote this matching relationship as $\Phi(\cdot)$. $V$ is a set that stores a large number of features (detailed in Sec. \ref{sec:cl}).

Besides, we also use the traditional binary cross-entropy loss as a basic optimization goal:
\vspace{-0.1cm}
\begin{equation}\label{eq:l_ce}
L_{c e}=-\sum_{i=1}^B \lambda_i\left(y_i \log p_i+\left(1-y_i\right) \log \left(1-p_i\right)\right)
\vspace{-0.18cm}
\end{equation}
where $p_i$ is the predicted score obtained by the binary classifier $h(\cdot)$. $\lambda_i$ is the weight consistent with Eq. (\ref{eq:l_guide}).

\begin{figure}[t]
\centering\includegraphics[width=0.47\textwidth]{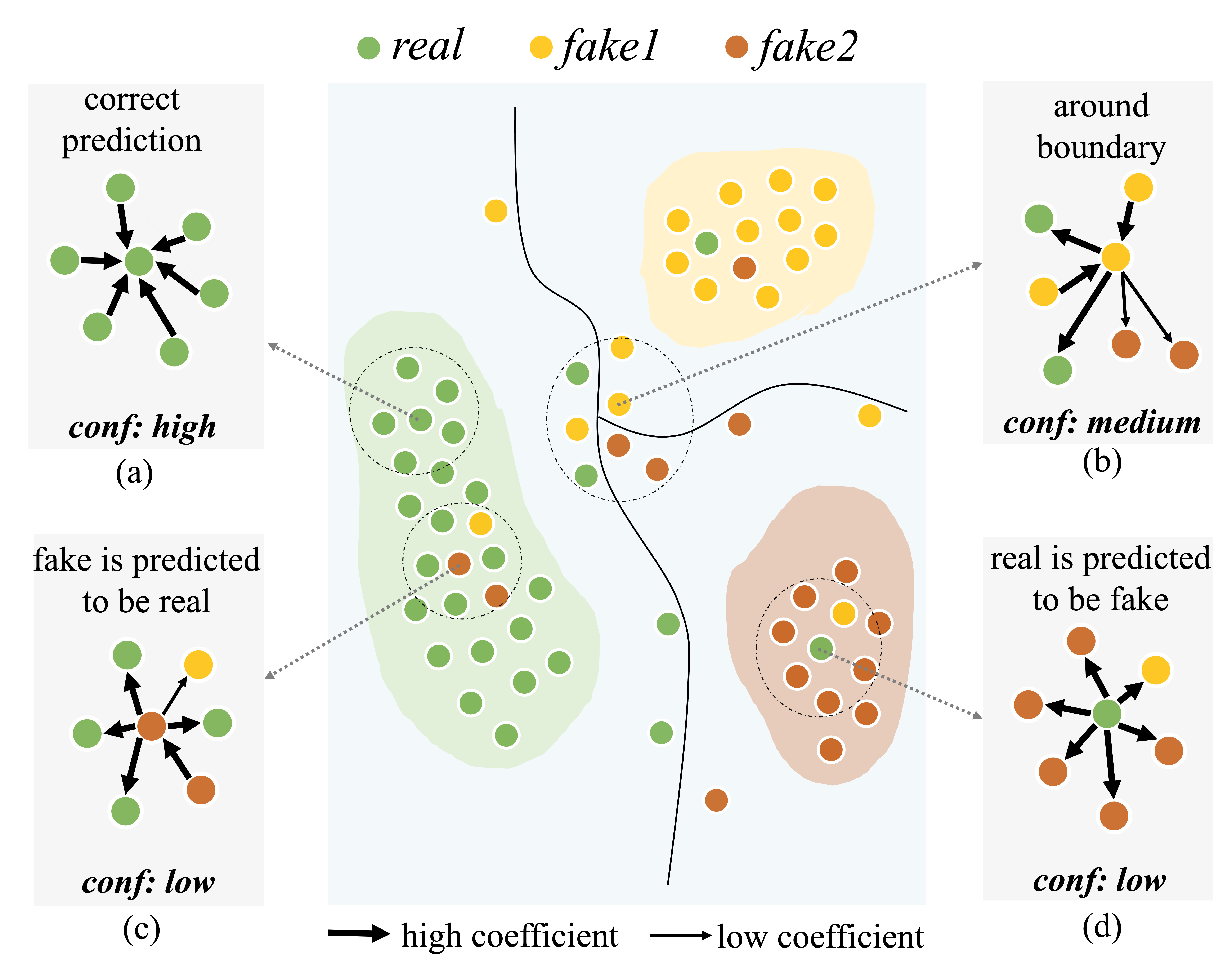}
\caption{Four cases in the calculation of the confidence. For each center point, inward arrows indicate that classes are consistent, and outward arrows indicate class inconsistencies. The thickness of the arrow reflects the value of the coefficient $\mu$ in Eq. (\ref{eq:conf}).}
\label{fig:conf}
\vspace{-0.08cm}
\end{figure}
\vspace{-0.1cm}
\subsection{Adjustment of decision boundary manifold}
\vspace{-0.1cm}
Furthermore, we design a module to adjust the decision boundary manifold (A-DBM). By focusing on some poorly performing samples within a batch, we strive to achieve the aggregation of features of the same domain, resulting in a better decision boundary manifold that conforms to the guidance of the guide-space.

Here, we define a metric called confidence to indicate the credibility of the model decision, which is calculated by the aggregation degree of the same domain features in the neighborhood of each feature point, and this process can be visualized in Figure \ref{fig:conf}.
For a sample whose neighbors in the feature space belong to the same domain (Figure \ref{fig:conf} (a)), the model prediction tends to be reliable and is assigned high confidence; while for the sample densely adjacent to other classes, it may be located in an area where another class is clustered, and tends to be mis-predicted as another class with high probability, corresponding to a low confidence (Figure \ref{fig:conf} (c) and (d)). For the points around the decision boundary, the model decision has uncertainty and the confidence is at the median (Figure \ref{fig:conf} (b)).

Specifically, for each image feature $v_i$, we calculate its similarity with each point in the feature set $V$, and then take the $k$ points with the highest similarity as its neighbors $K_i$ to obtain the adjacency relationship of the feature points.
According to the aggregation degree of the same class of features within the neighbors, the confidence $c_i$ of each data $x_i$ can be formulated as:
\vspace{-0.13cm}
\begin{equation}\label{eq:conf}
c_i=\frac{1}{\left|K_i\right|} \sum_{v_j \in K_i}\left(\mathbbm{1}_{t_i=t_j}-\mu \cdot \mathbbm{1}_{t_i \neq t_j}\right) \cdot \bm{\varepsilon}_{i j}
\vspace{-0.17cm}
\end{equation}
where $\bm{\varepsilon}_{i j}$ refers to the similarity between $v_i$ and $v_j$, and $\varepsilon_{i j}=\frac{1}{2}\left(1+v_i v_j\right)$.
$\mu$ is an adjustment coefficient. If $x_i$ is a real image, neighbors that are inconsistent with its class all belong to various fake categories with equal importance, i.e., $\mu=1$. However, if $x_i$ is a forgery image, the class inconsistency includes two situations of real and other forgery domains. It is more important to separate it from the real data than to separate it from data of other forgery domains, so $\mu=1$ for the former and $\mu=0.5$ for the latter.

Based on the confidence, we can find samples that perform poorly according to the aggregation degree of features of the same domain. Then we assign higher attention (i.e., higher weight $\lambda_i$ in Eq. (\ref{eq:l_guide}) and (\ref{eq:l_ce})) to these low-confidence samples to improve their clustering effect, resulting in a better decision boundary manifold.
Given confidence $c_i$ ($i \in[B]$) of data within a batch, $\lambda_i$ can be formulated as:
\begin{equation}\label{eq:lam}
\vspace{-0.1cm}
\lambda_i=\operatorname{softmax}\left(-c_i\right)
\vspace{-0.1cm}
\end{equation}

\vspace{-0.2cm}
\subsection{Decoupling module for irrelevant information} \label{sec:cl}
\vspace{-0.1cm}
In order for features to be more forgery-relevant, we design a decoupling module to alleviate the interference of irrelevant information on distinguishing different domains.
Before training, we first mine the potential correlation between images of the given training data.
We use a self-supervised model \cite{swinv2} pre-trained on face pictures to perform feature extraction on the training set. The self-supervised features have a certain degree of general representation ability for images.
Then we cluster the features \cite{cluster} to find correlations between the data, as shown in the decoupling module of Figure \ref{fig:frame}.
If features of different forgery domains are gathered into the same cluster, it proves that they have strong similarities in forgery-irrelevant information and need to be focused on in training to be separated. Conversely, if the features of the same domain belong to different clusters, they should be pulled closer to make features more related to forgery.

Let $\rho_i$ denote the cluster label of each instance $x_i$, we first construct a sample set $V_i^{+}$ that needs to be pulled closer and a sample set $V_i^{-}$ that needs to be pushed away. $V_i^{+}=\left\{v_j \in V \mid t_i=t_j, \rho_i \neq \rho_j\right\}$, $V_i^{-}=\left\{v_j \in V \mid t_i \neq t_j, \rho_i=\rho_j\right\}$.
We randomly select $n^+$ and $n^-$ features from $V_i^{+}$ and $V_i^{-}$ respectively to participate in the calculation, and the corresponding feature set can be denoted as 
$\tilde{V}_i^{+}$ and $\tilde{V}_i^{-}$ ($\tilde{V}_i^{+} \subset V_i^{+}$, $\tilde{V}_i^{-} \subset V_i^{-}$). The feature candidate set $V=\left\{\left(x_j, y_j, t_j\right)\right\}_{j=1}^Q$, which is a queue of dynamic accumulation of multiple batch features, with the current batch enqueued and the oldest batch dequeued. This $V$ with larger size $Q$ than the batch size allows better sampling in a broader and comprehensive selection scope. 

On the one hand, we push away samples in $\tilde{V}_i^{-}$ that belong to different domains from $x_i$ but are clustered by irrelevant information. This separation helps to reduce the irrelevant information contained in the features.
Based on KCL \cite{kcl} loss, the pushing loss is denoted as:
\vspace{-0.1cm}
\begin{equation}\label{eq:neg}
\small
L_{push}=\sum_{i=1}^B \lambda_i \cdot \frac{1}{1+n^{-}} \sum_{v_j^{-} \in \tilde{V}_i^{-}} \log \frac{e^{v_i^T v_j^{-} / \tau}}{\sum_{v_j \in V \cup G} e^{v_i^T v_j / \tau}}
\vspace{-0.1cm}
\end{equation}
\vspace{-0.1cm}

On the other hand, we pull the samples in $\tilde{V}_i^{+}$ closer.
This closeness due to the same domain rather than the similarity of irrelevant information implies an increase in the forgery-relevance of the features. The pulling loss is :
\vspace{-0.12cm}
\begin{equation}\label{eq:pos}
\small
L_{pull}=-\sum_{i=1}^B \lambda_i \cdot \frac{1}{1+n^{+}} \sum_{v_j^{+} \in \widetilde{V}_i^{+}} \log \frac{e^{v_i^T v_j^{+} / \tau}}{\sum_{v_j \in V \cup G} e^{v_i^T v_j / \tau}}
\vspace{-0.12cm}
\end{equation}

In summary, our overall loss can be formulated as:
\begin{equation}\label{eq:all}
L=\gamma_1 \cdot L_{guide}+\gamma_2 \cdot L_{c e}+\gamma_3 \cdot L_{pull}+\gamma_4 \cdot L_{push}
\end{equation}
where $\lambda_i$ is calculated by Eq. (\ref{eq:lam}), and $\gamma$ is the scale factor.


\vspace{-0.1cm}
\section{Experiments}
\vspace{-0.1cm}
\subsection{Experimental settings}\label{sec:expset}
\vspace{-0.1cm}
\paragraph{Datasets:}
We conduct experiments on three benchmark public forgery datasets:
1) FaceForensics++(FF++) \cite{ff++} contains four forgery methods (i.e., Deepfakes (DF) \cite{deepfake}, Face2Face (F2F) \cite{f2f}, FaceSwap (FS) \cite{fs}, and NeuralTextures (NT) \cite{nt}) with three image qualities including raw, high quality (HQ) and low quality (LQ).
2) CelebDF \cite{celebdf} contains real videos of 59 celebrities and corresponding high-quality fake videos generated by the improved forgery techniques.
3) Deepfake Detection Challenge (DFDC) \cite{dfdc} is a more challenging dataset that comes with the competition, with many manipulation and perturbation methods.
\vspace{-0.9cm}
\paragraph{Metrics:}  Following works in \cite{mat,recce,spsl,rfm,srm}, we use Accuracy score (Acc) and Area Under the Receiver Operating Characteristic Curve (AUC) as the metrics to evaluate the performance.
Between the two, we pay more attention to the AUC results, since the Acc values are affected by specific thresholds and data balance.
\vspace{-0.5cm}
\paragraph{Implementation Details:} In our experiments, we use EfficientNet-B4 (EN-B4) \cite{efb4} as the backbone when not otherwise specified. In the guide-space, $\theta_0=120^{\circ}$. 
For A-DBM, we adjust $\lambda_i$ from the 10-$th$ epoch, and before that, $\lambda_i=1/B$. $k=\left|K_i\right|=55$ in Eq. (\ref{eq:conf}).
In decoupling, the self-supervised model is trained under SimMIM \cite{simmim} framework, and the number of clusters is 500.
$n^+$ and $n^-$ are both 10. In Eq. (\ref{eq:all}), $\gamma_1=1$, $\gamma_2=0.5$, $\gamma_3=0.01$, and $\gamma_4=0.005$. The temperature parameter $\tau=1$. During the training, the batchsize $B=256$, and the size $Q$ of the set $V$ is 5120.
The maximum number of epochs is 60.
More details on hyper-parameters are shown in Appendix 3, and the analysis of computational cost introduced by the method is shown in Appendix 4.

\vspace{-0.25cm}
\subsection{In-Domain evaluations}\label{sec:indomain}
\vspace{-0.15cm}
We first verify the detection ability of our method against in-domain forgery methods (i.e., methods contained in the training set) on FF++ dataset. We train on both HQ and LQ image qualities using the four included forgery methods, and Table \ref{tab:indomain} lists the performance comparisons between ours and some current state-of-the-art methods.

\begin{table}[t]
\centering 
\resizebox{0.49\textwidth}{28.5mm}{
\begin{tabular}{l|p{0.9cm}<{\centering}p{0.9cm}<{\centering}|p{0.9cm}<{\centering}p{0.9cm}<{\centering}}
\toprule[1.2pt]
\multicolumn{1}{c|}{\multirow{2}{*}{Methods}} & \multicolumn{2}{c|}{FF++ (HQ)}  & \multicolumn{2}{c}{FF++ (LQ)}   \\ \cmidrule{2-5} 
\multicolumn{1}{c|}{}                         & Acc       & AUC        & Acc        & AUC       \\ \midrule[1pt]
Xception \cite{xception}                      & 95.04          & 96.30          & 84.11          & 92.50          \\ 
F$^3$-Net (Xception) \cite{f3net}           & 97.31          & 98.10          & 86.89          & 93.30          \\ 
EN-B4 \cite{efb4}                           & 96.63          & 99.18          & 86.67          & 88.20          \\ 
MAT (EN-B4) \cite{mat}                      & 97.60          & 99.29          & 88.69          & 90.40          \\ 
SPSL \cite{spsl}                            & 91.50          & 95.32          & 81.57          & 82.82          \\ 
RFM \cite{rfm}                              & 95.69          & 98.79          & 87.06          & 89.83          \\ 
Local-relation \cite{localrelation}         & 97.59          & 99.56          & 91.47          & 95.21          \\ 
RECCE \cite{recce}                          & 97.06          & 99.32          & 91.03          & 95.02          \\ 
CD-Net \cite{cdnet}                         & 98.75          & 99.90          & 88.12          & 95.20          \\ \midrule[1pt]
Ours                                        & \textbf{99.24} & \textbf{99.95} & \textbf{92.76} & \textbf{96.85} \\ \bottomrule[1.2pt]
\end{tabular}}
\vspace{0.01cm}
\caption{ In-domain comparisons on FF++ dataset. Results contain Acc (\%) and AUC (\%) of high quality (HQ) and low quality (LQ).}\label{tab:indomain}
\vspace{-0.2cm}
\end{table}

It can be seen that our method achieves the best performance on HQ and LQ with AUC of 99.95\% and 96.85\%, respectively, confirming that our method is effective for both high-quality and low-quality data.
On the HQ dataset, the AUC of our method is 1.16\% higher than that of RFM which enlarges the model's attention by erasing sensitive areas. F$^3$-Net \cite{f3net}, SPSL \cite{spsl}, and CD-Net \cite{cdnet} consider the frequency domain information, and ours are all ahead of them. 
On the LQ dataset, our AUC is 6.45\% higher than that of MAT \cite{mat} using the attention mechanism. Local-relation \cite{localrelation} fusing RGB and frequency domain information achieves the sub-optimal performance with the AUC of 95.21\%, while ours is still 1.64\% higher than that.

Although the above methods use attention, frequency information, local consistency, etc. to learn more generalized features, the supervision of binary classification makes the semantic and texture features still exist in the feature space.
In contrast, our method increases the discrimination between domains through the guide-space, and decouples forgery-irrelevant information from features, so as to learn more forgery-relevant representations.
\vspace{-0.2cm}
\subsection{Cross-domain evaluations for generalization}\label{sec:cross-domain}
\vspace{-0.12cm}
To verify the generalization of our method, we train the models using four forgery methods on the HQ dataset of FF++, and then test the cross-domain generalization on CelebDF and DFDC.
We compare with many state-of-the-art methods, such as DCL \cite{dcl} using the contrastive learning (CL), F$^3$-Net \cite{f3net}, SRM \cite{srm}, Local-relation \cite{localrelation} considering frequency domain information, UIA-ViT \cite{uia-vit} based on the transformer, LTW \cite{ltw} based on the meta-learning, and reconstruction-learning based RECCE \cite{recce}.
The corresponding AUC results are shown in Table \ref{tab:crossdomain}.

\begin{table}[t]
\centering 
\begin{tabular}{l|cc}
\toprule[1.2pt]
Methods                            & CelebDF        & DFDC           \\ \midrule[1pt]
Xception \cite{xception}           & 66.91          & 67.93          \\
F$^3$-Net (Xception) \cite{f3net}  & 71.21          & 72.88          \\
EN-B4 \cite{efb4}                  & 66.24          & 66.81          \\
MAT(EN-B4) \cite{mat}              & 76.65          & 67.34          \\
Face X-ray \cite{spsl}             & 74.20          & 70.00          \\
RFM \cite{rfm}                     & 67.64          & 68.01          \\
SRM \cite{srm}                     & 79.40          & 79.70          \\
Local-relation \cite{localrelation} & 78.26          & 76.53          \\
RECCE \cite{recce}                 & 77.39          & 76.75          \\
LTW \cite{ltw}                     & 77.14          & 74.58          \\
DCL \cite{dcl}                     & 82.30          & 76.71          \\
UIA-ViT \cite{uia-vit}             & 82.41          & 75.80          \\ \midrule[1pt]
Ours                               & \textbf{84.97} & \textbf{81.65} \\ \bottomrule[1.2pt]
\end{tabular}
\vspace{0.05cm}
\caption{Cross-domain comparisons of generalization based on AUC (\%). We train the detection model on the HQ dataset of FF++ and then test it on CelebDF and DFDC. }\label{tab:crossdomain}
\vspace{-0.18cm}
\end{table}


The AUC of our method on CelebDF and DFDC is 84.97\% and 81.65\%, respectively, outperforming other methods listed in Table \ref{tab:crossdomain}.
For CL-based methods, DCL focuses on the real-fake discrimination 
, achieving AUC of 82.30\% and 76.71\% on CelebDF and DFDC, while we further enhance the discrimination of different domains (real-fake, fake-fake), and AUCs are 2.67\% and 4.94\% higher than it, respectively.
For frequency-based methods, ours is 5.57\% ahead of SRM \cite{srm} on CelebDF.
For recent transformer-based UIA-ViT, our AUC leads by 2.56\% and 5.85\% on Celeb-DF and DFDC, respectively.
Although RECCE \cite{recce} also learns the common compact representations of real faces through reconstruction learning, its AUCs on the two datasets are 77.39\% and 76.75\%, which are lower than ours.
Unlike implicit learning in RECCE, we explicitly control how compact the real representation is by controlling the angle $\theta_0$ between the real and forgery embeddings in the guide-space. So coupled with the efforts that we also focus on the discrimination between different forgery domains to capture more forgery-related traces, we can achieve better performance.

\begin{table*}[t]
\centering 
\resizebox{0.98\textwidth}{30mm}{
\begin{tabular}{p{2.5cm}|p{0.9cm}<{\centering}p{0.9cm}<{\centering}p{0.9cm}<{\centering}p{0.9cm}<{\centering}p{0.9cm}<{\centering}p{0.9cm}<{\centering}p{0.9cm}<{\centering}p{0.9cm}<{\centering}|p{0.9cm}<{\centering}p{0.9cm}<{\centering}p{0.9cm}<{\centering}p{0.9cm}<{\centering}}
\toprule[1.2pt]
Train   Set  & \multicolumn{4}{c}{F2F   FS NT}                                   & \multicolumn{4}{c|}{DF FS NT}                                    & \multicolumn{4}{c}{FF++   (HQ)}                                   \\ \cmidrule(r){1-1} \cmidrule(r){2-5} \cmidrule(r){6-9} \cmidrule(r){10-13}
Test Set     & \multicolumn{2}{c}{DF (HQ)}     & \multicolumn{2}{c}{DF (LQ)}     & \multicolumn{2}{c}{F2F (HQ)}     & \multicolumn{2}{c|}{F2F (LQ)}    & \multicolumn{2}{c}{CelebDF}     & \multicolumn{2}{c}{DFDC}        \\ \cmidrule(r){1-1} \cmidrule(r){2-3} \cmidrule(r){4-5} \cmidrule(r){6-7} \cmidrule(r){8-9} \cmidrule(r){10-11} \cmidrule(r){12-13}
             & Acc            & AUC            & Acc            & AUC            & Acc            & AUC            & Acc            & AUC            & Acc            & AUC            & Acc            & AUC            \\ \cmidrule(r){2-3} \cmidrule(r){4-5} \cmidrule(r){6-7} \cmidrule(r){8-9} \cmidrule(r){10-11} \cmidrule(r){12-13}
$L_{ce-2}$         & 91.97          & 92.48          & 91.46          & 96.71          & 84.44          & 91.17          & \textbf{91.37} & 96.49          & 62.93          & 66.24          & 62.16          & 66.81                   \\
$L_{ce-(1+N)}$     & 93.83          & 95.59          & 92.34          & 97.32          & 83.91          & 91.15          & 91.08          & 96.42          & 65.06          & 67.74          & 65.16          & 68.58                   \\
$L_{guide}$        & \textbf{94.21} & \textbf{95.76} & \textbf{93.96} & \textbf{97.79} & \textbf{85.54} & \textbf{92.33} & 90.42          & \textbf{96.87} & \textbf{66.02} & \textbf{70.72} & \textbf{67.67} & \textbf{71.47}          \\ \midrule[0.7pt]
w/o $L_{guide}$    & 93.07          & 96.91          & 93.62          & 97.64          & 87.85          & 93.56          & 95.72          & 98.67          & 68.59          & 76.42          & 69.37          & 75.12                   \\
w/o $L_{pull}$\&$L_{push}$ & 95.82    & 98.15          & 95.73          & 98.59          & 90.26          & 94.19          & 95.90          & 99.08          & 71.52          & 79.13          & 72.85          & 78.04                  \\
w/o $L_{pull}$      & 97.12          & 98.87          & 96.62         & 99.11           & 92.65          & 95.82          & 96.17          & 99.32          & 71.63          & 80.25          & 73.09          & 78.98                  \\
w/o $L_{push}$      & 97.46          & 99.39          & 97.71         & 99.57           & 93.14          & 97.50          & 96.91          & 99.49          & 71.94          & 81.78          & 73.55          & 79.26                 \\
w/o {\small A-DBM} & 96.03          & 97.40          & 96.47         & 98.79           & 91.28          & 96.35          & 95.87          & 99.02          & 72.12          & 78.92          & 70.23          & 77.15              \\ \midrule[0.6pt]
ours         & \textbf{98.92}       & \textbf{99.81}  & \textbf{98.72}  & \textbf{99.89}  & \textbf{95.76}  & \textbf{98.92}  & \textbf{97.96} & \textbf{99.68}  & \textbf{73.19} & \textbf{84.97}  & \textbf{74.83}   & \textbf{81.65}    \\ \bottomrule[1.2pt]
\end{tabular}}
\vspace{0.08cm}
\caption{Ablation results in two settings: 1) Cross-test within FF++ (left); 2) From FF++ to others (right). The upper part compares the three losses that increase the discrimination of different domains; the lower part shows the performance after removing each module of the method.
}\label{tab:abla}
\vspace{-0.1cm}
\end{table*}
\vspace{-0.14cm}
\subsection{Ablation study}\label{sec:abla}
\vspace{-0.14cm}
In this section, we conduct detailed studies of each module involved in the method. The evaluation of generalization follows two settings: 1) Cross-test within FF++: training with three methods within FF++ and testing on the remaining one; 2) From FF++ to others: training on the four forgery methods of FF++ and testing on CelebDF and DFDC.
\vspace{-0.5cm}
\paragraph{Methods to distinguish different forgery domains:} We first analyze methods of enhancing the forgery domain discrimination to verify our superiority based on the guide-space.
The training result based on binary cross-entropy ($L_{ce-2}$) loss is an experimental baseline.
On this basis, we compare the performance of using multi-class cross-entropy ($L_{ce-(1+N)}$) loss, where the number of classes is the number of forgery domains $N$ plus 1.
The performance comparisons of $L_{ce-2}$, $L_{ce-(1+N)}$, and our $L_{guide}$ under two experimental settings are shown in the upper part of Table \ref{tab:abla}.
For the cross-test within FF++, we experiment on both HQ and LQ datasets. Due to space limitations, we here show the results of DF and F2F as the test set, and the remaining results of FS and NT are shown in Appendix 5.1.

It can be seen that the AUC of our guide-method $L_{guide}$ achieves optimal performance among the three losses at all of these settings. Especially on CelebDF and DFDC, AUCs are 4.48\% and 4.66\% higher than $L_{ce-2}$, and 2.98\% and 2.89\% higher than $L_{ce-(1+N)}$.
For $L_{ce-(1+N)}$, it has improved performance over $L_{ce-2}$ in most cases,
but in some cases, it is the opposite, for example, Acc and AUC on F2F(HQ) are 0.53\% and 0.02\% lower than $L_{ce-2}$.
This shows that simply uncontrolled separation of several forgery domains in $L_{ce-(1+N)}$ is sometimes not feasible, because it treats the real and forgery domain as equal classes, so that the distances between real-forgery and between forgery-forgery are equal.
However, we expect the separation of real-forgery to be much greater than forgery-forgery to achieve the compactness of the real domain representation.
Our guide method can achieve this by actively controlling the angle between the real and forgery guide embeddings, so it can achieve the optimal performance.

\vspace{-0.6cm}
\paragraph{Importance of different modules:} We mainly study the guide embedding ($L_{guide}$), the decoupling module ($L_{pull}$, $L_{push}$), and the
boundary adjustment (A-DBM) included in the method, and examine the importance by removing the corresponding module from the overall method. The results are shown in the lower half of Table \ref{tab:abla}, and the remaining results of FS and NT are shown in Appendix 5.2.


\begin{table}[t]
\centering 
\resizebox{0.42\textwidth}{25mm}{
\begin{tabular}{p{2.5cm}|p{0.85cm}<{\centering}p{0.85cm}<{\centering}p{0.85cm}<{\centering}p{0.85cm}<{\centering}}
\toprule[1.2pt]
\multirow{2}{*}{Backbone} & \multicolumn{2}{c}{CelebDF} & \multicolumn{2}{c}{DFDC} \\ \cmidrule[0.6pt]{2-5}
                          & Acc          & AUC          & Acc         & AUC        \\ \midrule[0.8pt]
Xception                  & 64.09        & 66.91        & 62.16       & 67.93      \\
Xception+Ours             & 69.31        & 76.46        & 66.29       & 75.21      \\ \midrule[0.6pt]
Resnet50                  & 62.47        & 67.08        & 64.16       & 67.68      \\
Resnet50+Ours             & 71.92        & 77.05        & 70.69       & 74.03      \\ \midrule[0.6pt]
DPN68                     & 64.86        & 70.78        & 64.16       & 67.72      \\
DPN68+Ours                & 72.73        & 80.09        & 68.98       & 77.25      \\ \midrule[0.6pt]
VGG19                     & 67.18        & 71.31        & 68.17       & 72.56      \\
VGG19+Ours                & 74.91        & 81.89        & 71.54       & 78.75      \\ \bottomrule[1.2pt]
\end{tabular}}
\vspace{0.05cm}
\caption{Generalization when using other backbones. Models are trained on FF++(HQ) and tested on CelebDF and DFDC.}\label{tab:backbone}
\vspace{-0.3cm}
\end{table}

Overall, removing the guide method (w/o $L_{guide}$) has the greatest impact, e.g., the AUC on CelebDF is reduced from 84.97\% to 76.42\%. Although the decoupling module can also achieve the separation of different domains and the closeness of the same domain, it lacks predefined guide vectors and cannot actively control the degree of separation between domains.
The performance also drops significantly when the decoupling module is removed (w/o $L_{pull}$\&$L_{push}$), demonstrating the importance of alleviating the association of irrelevant information for features.
Based on the results, removing the pulling set (w/o $L_{pull}$) has a greater negative impact than removing the pushing set (w/o $L_{push}$), for example, on CelebDF, AUCs decrease by 4.72\% and 3.19\%, respectively.
The boundary module works well in all experimental settings, especially with 6.05\% and 4.5\% increase of AUC on CelebDF and DFDC after using it.
In addition, under the left setting (1) of Table \ref{tab:abla}, the performance comparison of our method with other SOTA methods is shown in Appendix 5.3.

\vspace{-0.5cm}
\paragraph{Generalization when using other backbones:} Besides EfficientNet-B4 \cite{efb4}, our method can also be used in other backbone networks. Table \ref{tab:backbone} lists the performance of our method under Xception \cite{xception}, Resnet50 \cite{resnet}, DPN68 \cite{dpn} and VGG19 \cite{vgg}.
The models are trained on FF++ (HQ) and tested on CelebDF and DFDC.
Compared with the original training, the performance on two datasets has improved after using our training method, e.g., AUCs of the four models on CelebDF are improved by 9.55\%, 9.97\%, 9.31\%, and 10.58\%, respectively.
This demonstrates that our method has good adaptability and can be combined with various backbone networks to achieve better performance.

\subsection{Analysis and visualization}\label{sec:analysis}
\vspace{-0.1cm}
\paragraph{Predefined $\theta_0$ in guide-space:} When constructing the guide-space, we pre-define the angle $\theta_0$ of real-forgery domains, and maximize the forgery-forgery domain separation $\theta_{i j}(1 \leq i, j \leq N)$.  $\theta_{i j}$ can be equal when $d \geq N$.
However, in the limited feature space, $\theta_0$ and $\theta_{i j}$ have a trade-off, i.e., a large $\theta_0$ means that the corresponding $\theta_{i j}$ will be small.
We study the effect of different $\theta_0$, and the corresponding $\theta_{i j}$ and the generalization results are listed in Table \ref{tab:angle}.

\begin{table}[t]
\centering 
\resizebox{0.44\textwidth}{21mm}{
\begin{tabular}{p{0.97cm}<{\centering}|p{0.97cm}<{\centering}|p{0.96cm}<{\centering}p{0.96cm}<{\centering}p{0.96cm}<{\centering}p{0.96cm}<{\centering}}
\toprule[1.2pt]
\multirow{2}{*}{$\theta_0$} & \multirow{2}{*}{$\theta_{ij}$} & \multicolumn{2}{c}{CelebDF} & \multicolumn{2}{c}{DFDC} \\\cmidrule[0.6pt]{3-6}
                                          &                                                & Acc          & AUC          & Acc         & AUC        \\ \midrule[0.6pt]
$90^\circ$                                & $109^\circ$                                    & 69.22        & 74.95        & 66.34       & 74.82      \\
$100^\circ$                               & $107^\circ$                                    & 70.19        & 77.86        & 71.74       & 76.95      \\
$110^\circ$                               & $100^\circ$                                    & 72.58        & 81.07        & 73.91       & 79.23      \\
$120^\circ$                               & $90^\circ$                                     & \textbf{73.19}  & \textbf{84.97}  & \textbf{74.83} & \textbf{81.65}      \\
$130^\circ$                               & $78^\circ$                                     & 71.32        & 80.83        & 73.56       & 80.19      \\
$140^\circ$                               & $63^\circ$                                     & 70.78        & 79.05        & 71.29       & 78.46      \\
$150^\circ$                               & $48^\circ$                                     & 70.20        & 78.52        & 70.96       & 77.87      \\ \bottomrule[1.2pt]
\end{tabular}}
\vspace{0.05cm}
\caption{The value of $\theta_{i j}(1 \leq i, j \leq N)$ and the corresponding generalization performance when $\theta_0$ takes different values.}\label{tab:angle}
\vspace{-0.15cm}
\end{table}

Ideally, we expect $\theta_0>\theta_{i j}$, and $\theta_{i j}$ is as large as possible. In Table \ref{tab:angle}, the optimum performance is achieved when $\theta_0=120^\circ$, and the corresponding $\theta_{i j}=90^\circ$. $\theta_0$ is larger than $\theta_{i j}$, and the forgery-forgery is orthogonal, realizing the separation of each other.
When we decrease $\theta_0$ to $90^\circ$, $\theta_{i j}$ increases to $109^\circ$, and the performance drops significantly, e.g., the AUC of CelebDF decreases from 84.97\% to 74.95\%. At this time, although the forgery-forgery distinction is enhanced, the real and forgery domains are too close, which will limit the generalization.
Conversely, when we increase $\theta_0$ to $150^\circ$ , the real-forgery distance is increased, but the forgery-forgery $\theta_{i j}$ is reduced to $48^\circ$. As a result, the discrimination between forgery domains is weakened, and the AUC on CelebDF drops to 78.52\%.

\vspace{-0.5cm}
\paragraph{Visualizations of t-sne:} In Figure \ref{fig:tsne}, we visualize the feature space using t-sne \cite{tsne}, comparing the effect of binary cross-entropy (CE-2) with our method. For CE-2, the points of the forgery domains are messily mixed together without being distinguished by forgery domains. Instead, the features of ours are clustered according to the domain type, indicating that the learned features are more forgery-related.
\begin{figure}[t]
\centering\includegraphics[width=0.45\textwidth]{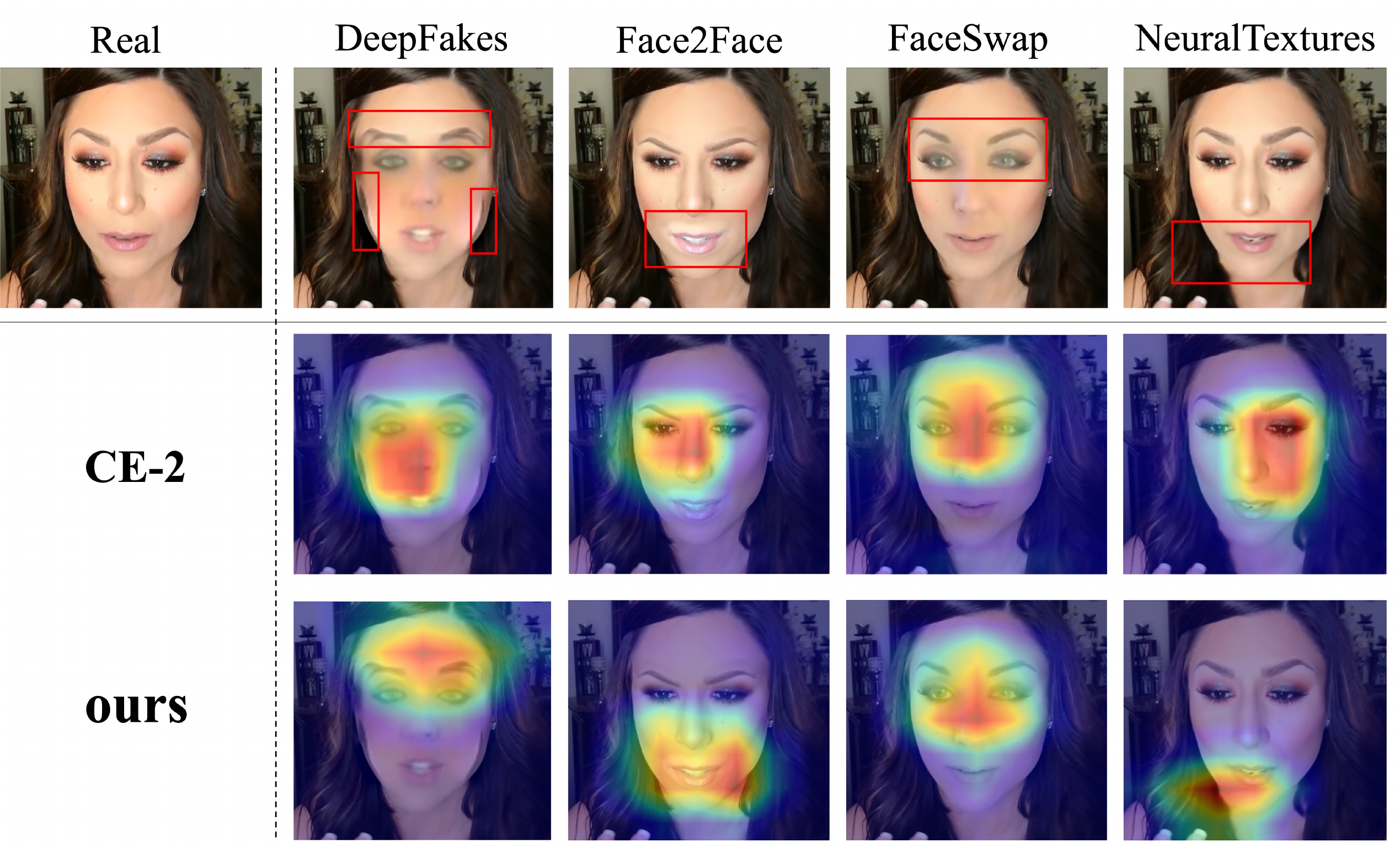}
\caption{The heatmap comparisons of binary cross-entropy (CE-2) and our method. Forgery artifacts are marked in red frames.}
\label{fig:heatmap}
\end{figure}
\begin{figure}[t]
\centering\includegraphics[width=0.45\textwidth]{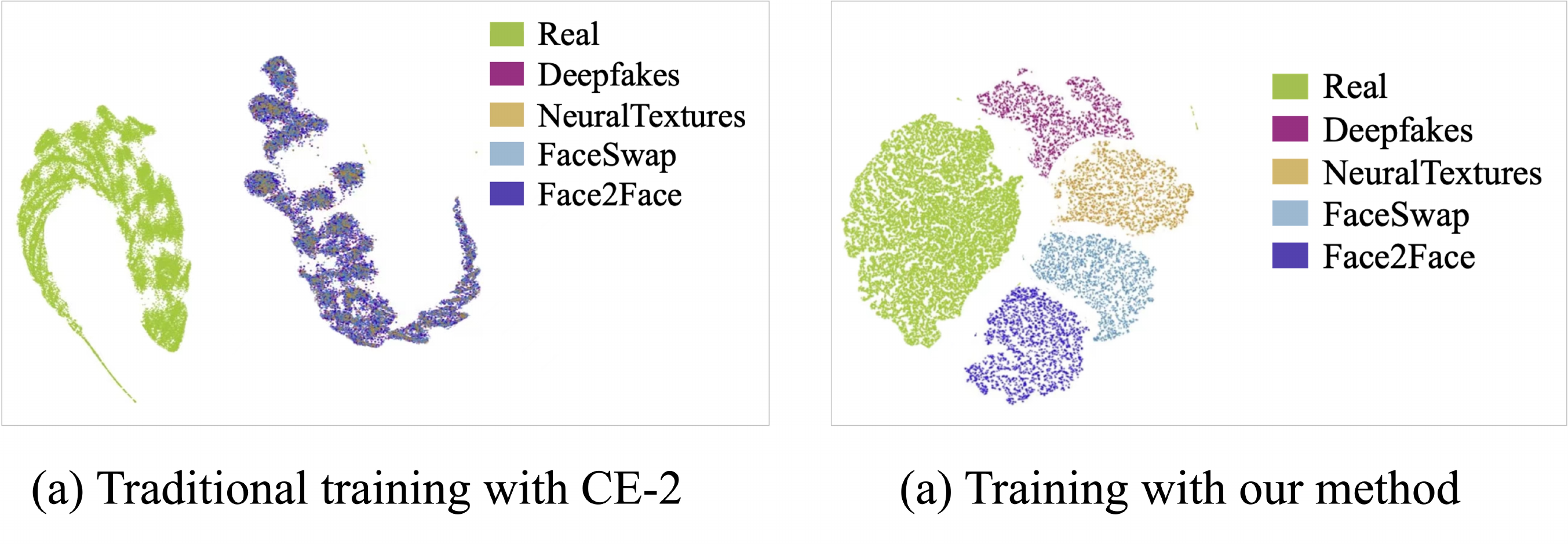}
\caption{The t-sne comparisons of binary cross-entropy (CE-2) and our method.}
\label{fig:tsne}
\vspace{-0.15cm}
\end{figure}
\vspace{-0.9cm}
\paragraph{Visualizations of heatmaps:} Identity is one of the forgery-irrelevant similarities existed between different domains.
In Figure \ref{fig:heatmap}, we compare the heatmaps for images of different forgery domains with the same identity when using CE-2 and our method.
When using CE-2, different domains are treated as the same fake, resulting in the model focusing on similar regions.
As shown in the second line of Figure \ref{fig:heatmap}, the concern areas are concentrated in the middle of the face rather than their respective forgery traces, indicating that the model is still perturbed by similarities of the same identity.
However, in the third line, our method focuses on respective forgery traces of different forgeries, which shows that our method mitigates the interference of these similarities and learns better forgery-related information. More heatmaps are shown in Appendix 6.

\vspace{-0.1cm}
\section{Conclusion}
\vspace{-0.1cm}
In this paper, we proposed a novel guide-space based framework to improve the generalization of face forgery detectors.
The well-designed guide-space can achieve separations of both real-forgery domains and forgery-forgery domains in a controllable manner, so as to capture more forgery-related information and ensure a large distance between real and fake representations simultaneously.
Furthermore, we used a decoupling module to reduce the interference of forgery-irrelevant inter-domain correlations for domain discrimination.
In addition, we designed a decision boundary adjustment module to make the features better follow the guidance of the guide-space. Extensive cross-domain experiments demonstrate the better generalization of our method.


{\small
\bibliographystyle{ieee_fullname}
\bibliography{main}

\begin{thebibliography}{10}\itemsep=-1pt

\bibitem{deepfake}
Deepfakes.
\newblock \url{https://github.com/deepfakes/faceswap}.
\newblock Accessed 2022-10-29.

\bibitem{fs}
Faceswap.
\newblock \url{https://github.com/MarekKowalski/FaceSwap}.
\newblock Accessed 2022-10-29.

\bibitem{mesonet}
Darius Afchar, Vincent Nozick, Junichi Yamagishi, and Isao Echizen.
\newblock Mesonet: a compact facial video forgery detection network.
\newblock In {\em IEEE international workshop on information forensics and
  security (WIFS)}, pages 1--7. IEEE, 2018.

\bibitem{lagrangian}
Brian Beavis and Ian Dobbs.
\newblock {\em Optimisation and stability theory for economic analysis}.
\newblock Cambridge university press, 1990.

\bibitem{recce}
Junyi Cao, Chao Ma, Taiping Yao, Shen Chen, Shouhong Ding, and Xiaokang Yang.
\newblock End-to-end reconstruction-classification learning for face forgery
  detection.
\newblock In {\em Proceedings of the IEEE/CVF Conference on Computer Vision and
  Pattern Recognition}, pages 4113--4122, 2022.

\bibitem{advtrain_cvpr22}
Liang Chen, Yong Zhang, Yibing Song, Lingqiao Liu, and Jue Wang.
\newblock Self-supervised learning of adversarial example: Towards good
  generalizations for deepfake detection.
\newblock In {\em Proceedings of the IEEE/CVF Conference on Computer Vision and
  Pattern Recognition}, pages 18710--18719, 2022.

\bibitem{localrelation}
Shen Chen, Taiping Yao, Yang Chen, Shouhong Ding, Jilin Li, and Rongrong Ji.
\newblock Local relation learning for face forgery detection.
\newblock In {\em Proceedings of the AAAI Conference on Artificial
  Intelligence}, volume~35, pages 1081--1088, 2021.

\bibitem{dpn}
Yunpeng Chen, Jianan Li, Huaxin Xiao, Xiaojie Jin, Shuicheng Yan, and Jiashi
  Feng.
\newblock Dual path networks.
\newblock {\em Advances in neural information processing systems}, 30, 2017.

\bibitem{cluster}
Yingjie Chen, Huasong Zhong, Chong Chen, Chen Shen, Jianqiang Huang, Tao Wang,
  Yun Liang, and Qianru Sun.
\newblock On mitigating hard clusters for face clustering.
\newblock In {\em Computer Vision--ECCV 2022: 17th European Conference, Tel
  Aviv, Israel, October 23--27, 2022, Proceedings, Part XII}, pages 529--544.
  Springer, 2022.

\bibitem{attention1}
Hao Dang, Feng Liu, Joel Stehouwer, Xiaoming Liu, and Anil~K Jain.
\newblock On the detection of digital face manipulation.
\newblock In {\em Proceedings of the IEEE/CVF Conference on Computer Vision and
  Pattern recognition}, pages 5781--5790, 2020.

\bibitem{dfdc}
Brian Dolhansky, Joanna Bitton, Ben Pflaum, Jikuo Lu, Russ Howes, Menglin Wang,
  and Cristian~Canton Ferrer.
\newblock The deepfake detection challenge (dfdc) dataset.
\newblock {\em arXiv preprint arXiv:2006.07397}, 2020.

\bibitem{sola}
Jianwei Fei, Yunshu Dai, Peipeng Yu, Tianrun Shen, Zhihua Xia, and Jian Weng.
\newblock Learning second order local anomaly for general face forgery
  detection.
\newblock In {\em Proceedings of the IEEE/CVF Conference on Computer Vision and
  Pattern Recognition}, pages 20270--20280, 2022.

\bibitem{infoswap}
Gege Gao, Huaibo Huang, Chaoyou Fu, Zhaoyang Li, and Ran He.
\newblock Information bottleneck disentanglement for identity swapping.
\newblock In {\em Proceedings of the IEEE/CVF conference on computer vision and
  pattern recognition}, pages 3404--3413, 2021.

\bibitem{lips}
Alexandros Haliassos, Konstantinos Vougioukas, Stavros Petridis, and Maja
  Pantic.
\newblock Lips don't lie: A generalisable and robust approach to face forgery
  detection.
\newblock In {\em Proceedings of the IEEE/CVF conference on computer vision and
  pattern recognition}, pages 5039--5049, 2021.

\bibitem{resnet}
Kaiming He, Xiangyu Zhang, Shaoqing Ren, and Jian Sun.
\newblock Deep residual learning for image recognition.
\newblock In {\em Proceedings of the IEEE conference on computer vision and
  pattern recognition}, pages 770--778, 2016.

\bibitem{shapes}
Katherine Hermann and Andrew Lampinen.
\newblock What shapes feature representations? exploring datasets,
  architectures, and training.
\newblock {\em Advances in Neural Information Processing Systems},
  33:9995--10006, 2020.

\bibitem{frepgan}
Yonghyun Jeong, Doyeon Kim, Youngmin Ro, and Jongwon Choi.
\newblock Frepgan: Robust deepfake detection using frequency-level
  perturbations.
\newblock 2022.

\bibitem{kcl}
Bingyi Kang, Yu Li, Sa Xie, Zehuan Yuan, and Jiashi Feng.
\newblock Exploring balanced feature spaces for representation learning.
\newblock In {\em International Conference on Learning Representations}, 2020.

\bibitem{fretal}
Minha Kim, Shahroz Tariq, and Simon~S Woo.
\newblock Fretal: Generalizing deepfake detection using knowledge distillation
  and representation learning.
\newblock In {\em Proceedings of the IEEE/CVF conference on computer vision and
  pattern recognition}, pages 1001--1012, 2021.

\bibitem{hungarian}
Harold~W Kuhn.
\newblock The hungarian method for the assignment problem.
\newblock {\em Naval research logistics quarterly}, 2(1-2):83--97, 1955.

\bibitem{scl}
Jiaming Li, Hongtao Xie, Jiahong Li, Zhongyuan Wang, and Yongdong Zhang.
\newblock Frequency-aware discriminative feature learning supervised by
  single-center loss for face forgery detection.
\newblock In {\em Proceedings of the IEEE/CVF conference on computer vision and
  pattern recognition}, pages 6458--6467, 2021.

\bibitem{facexray}
Lingzhi Li, Jianmin Bao, Ting Zhang, Hao Yang, Dong Chen, Fang Wen, and Baining
  Guo.
\newblock Face x-ray for more general face forgery detection.
\newblock In {\em Proceedings of the IEEE/CVF conference on computer vision and
  pattern recognition}, pages 5001--5010, 2020.

\bibitem{li2022learning}
Yinghui Li, Chen Wang, Li Yangning, Hai-Tao Zheng, and Ying Shen.
\newblock Learning purified feature representations from task-irrelevant
  labels.
\newblock In {\em 2022 International Joint Conference on Neural Networks
  (IJCNN)}, pages 01--08. IEEE, 2022.

\bibitem{celebdf}
Yuezun Li, Xin Yang, Pu Sun, Honggang Qi, and Siwei Lyu.
\newblock Celeb-df: A large-scale challenging dataset for deepfake forensics.
\newblock In {\em Proceedings of the IEEE/CVF conference on computer vision and
  pattern recognition}, pages 3207--3216, 2020.

\bibitem{spsl}
Honggu Liu, Xiaodan Li, Wenbo Zhou, Yuefeng Chen, Yuan He, Hui Xue, Weiming
  Zhang, and Nenghai Yu.
\newblock Spatial-phase shallow learning: rethinking face forgery detection in
  frequency domain.
\newblock In {\em Proceedings of the IEEE/CVF conference on computer vision and
  pattern recognition}, pages 772--781, 2021.

\bibitem{swinv2}
Ze Liu, Han Hu, Yutong Lin, Zhuliang Yao, Zhenda Xie, Yixuan Wei, Jia Ning, Yue
  Cao, Zheng Zhang, Li Dong, Furu Wei, and Baining Guo.
\newblock Swin transformer v2: Scaling up capacity and resolution.
\newblock In {\em International Conference on Computer Vision and Pattern
  Recognition (CVPR)}, 2022.

\bibitem{srm}
Yuchen Luo, Yong Zhang, Junchi Yan, and Wei Liu.
\newblock Generalizing face forgery detection with high-frequency features.
\newblock In {\em Proceedings of the IEEE/CVF conference on computer vision and
  pattern recognition}, pages 16317--16326, 2021.

\bibitem{twobranch}
Iacopo Masi, Aditya Killekar, Royston~Marian Mascarenhas, Shenoy~Pratik
  Gurudatt, and Wael AbdAlmageed.
\newblock Two-branch recurrent network for isolating deepfakes in videos.
\newblock In {\em European conference on computer vision}, pages 667--684.
  Springer, 2020.

\bibitem{rl_aug}
Aakash~Varma Nadimpalli and Ajita Rattani.
\newblock On improving cross-dataset generalization of deepfake detectors.
\newblock In {\em Proceedings of the IEEE/CVF Conference on Computer Vision and
  Pattern Recognition}, pages 91--99, 2022.

\bibitem{multitask}
Huy~H Nguyen, Fuming Fang, Junichi Yamagishi, and Isao Echizen.
\newblock Multi-task learning for detecting and segmenting manipulated facial
  images and videos.
\newblock In {\em 2019 IEEE 10th International Conference on Biometrics Theory,
  Applications and Systems (BTAS)}, pages 1--8. IEEE, 2019.

\bibitem{capsule}
Huy~H Nguyen, Junichi Yamagishi, and Isao Echizen.
\newblock Capsule-forensics: Using capsule networks to detect forged images and
  videos.
\newblock In {\em ICASSP 2019-2019 IEEE International Conference on Acoustics,
  Speech and Signal Processing (ICASSP)}, pages 2307--2311. IEEE, 2019.

\bibitem{f3net}
Yuyang Qian, Guojun Yin, Lu Sheng, Zixuan Chen, and Jing Shao.
\newblock Thinking in frequency: Face forgery detection by mining
  frequency-aware clues.
\newblock In {\em European conference on computer vision}, pages 86--103.
  Springer, 2020.

\bibitem{xception}
Andreas Rossler, Davide Cozzolino, Luisa Verdoliva, Christian Riess, Justus
  Thies, and Matthias Nie{\ss}ner.
\newblock Faceforensics++: Learning to detect manipulated facial images.
\newblock In {\em Proceedings of the IEEE/CVF international conference on
  computer vision}, pages 1--11, 2019.

\bibitem{ff++}
Andreas Rossler, Davide Cozzolino, Luisa Verdoliva, Christian Riess, Justus
  Thies, and Matthias Nie{\ss}ner.
\newblock Faceforensics++: Learning to detect manipulated facial images.
\newblock In {\em Proceedings of the IEEE/CVF international conference on
  computer vision}, pages 1--11, 2019.

\bibitem{selfblended}
Kaede Shiohara and Toshihiko Yamasaki.
\newblock Detecting deepfakes with self-blended images.
\newblock In {\em Proceedings of the IEEE/CVF Conference on Computer Vision and
  Pattern Recognition}, pages 18720--18729, 2022.

\bibitem{vgg}
Karen Simonyan and Andrew Zisserman.
\newblock Very deep convolutional networks for large-scale image recognition.
\newblock {\em arXiv preprint arXiv:1409.1556}, 2014.

\bibitem{cdnet}
Luchuan Song, Zheng Fang, Xiaodan Li, Xiaoyi Dong, Zhenchao Jin, Yuefeng Chen,
  and Siwei Lyu.
\newblock Adaptive face forgery detection in cross domain.
\newblock In {\em European Conference on Computer Vision}, pages 467--484.
  Springer, 2022.

\bibitem{ltw}
Ke Sun, Hong Liu, Qixiang Ye, Yue Gao, Jianzhuang Liu, Ling Shao, and Rongrong
  Ji.
\newblock Domain general face forgery detection by learning to weight.
\newblock In {\em Proceedings of the AAAI conference on artificial
  intelligence}, volume~35, pages 2638--2646, 2021.

\bibitem{dcl}
Ke Sun, Taiping Yao, Shen Chen, Shouhong Ding, Jilin Li, and Rongrong Ji.
\newblock Dual contrastive learning for general face forgery detection.
\newblock In {\em Proceedings of the AAAI Conference on Artificial
  Intelligence}, volume~36, pages 2316--2324, 2022.

\bibitem{geometry}
Zekun Sun, Yujie Han, Zeyu Hua, Na Ruan, and Weijia Jia.
\newblock Improving the efficiency and robustness of deepfakes detection
  through precise geometric features.
\newblock In {\em Proceedings of the IEEE/CVF Conference on Computer Vision and
  Pattern Recognition}, pages 3609--3618, 2021.

\bibitem{efb4}
Mingxing Tan and Quoc Le.
\newblock Efficientnet: Rethinking model scaling for convolutional neural
  networks.
\newblock In {\em International conference on machine learning}, pages
  6105--6114. PMLR, 2019.

\bibitem{nt}
Justus Thies, Michael Zollh{\"o}fer, and Matthias Nie{\ss}ner.
\newblock Deferred neural rendering: Image synthesis using neural textures.
\newblock {\em ACM Transactions on Graphics (TOG)}, 38(4):1--12, 2019.

\bibitem{f2f}
Justus Thies, Michael Zollhofer, Marc Stamminger, Christian Theobalt, and
  Matthias Nie{\ss}ner.
\newblock Face2face: Real-time face capture and reenactment of rgb videos.
\newblock In {\em Proceedings of the IEEE conference on computer vision and
  pattern recognition}, pages 2387--2395, 2016.

\bibitem{tsne}
Laurens Van~der Maaten and Geoffrey Hinton.
\newblock Visualizing data using t-sne.
\newblock {\em Journal of machine learning research}, 9(11), 2008.

\bibitem{rfm}
Chengrui Wang and Weihong Deng.
\newblock Representative forgery mining for fake face detection.
\newblock In {\em Proceedings of the IEEE/CVF conference on computer vision and
  pattern recognition}, pages 14923--14932, 2021.

\bibitem{osn}
Haiwei Wu, Jiantao Zhou, Jinyu Tian, and Jun Liu.
\newblock Robust image forgery detection over online social network shared
  images.
\newblock In {\em Proceedings of the IEEE/CVF Conference on Computer Vision and
  Pattern Recognition}, pages 13440--13449, 2022.

\bibitem{simmim}
Zhenda Xie, Zheng Zhang, Yue Cao, Yutong Lin, Jianmin Bao, Zhuliang Yao, Qi
  Dai, and Han Hu.
\newblock Simmim: A simple framework for masked image modeling.
\newblock In {\em International Conference on Computer Vision and Pattern
  Recognition (CVPR)}, 2022.

\bibitem{mat}
Hanqing Zhao, Wenbo Zhou, Dongdong Chen, Tianyi Wei, Weiming Zhang, and Nenghai
  Yu.
\newblock Multi-attentional deepfake detection.
\newblock In {\em Proceedings of the IEEE/CVF conference on computer vision and
  pattern recognition}, pages 2185--2194, 2021.

\bibitem{PCLI2G}
Tianchen Zhao, Xiang Xu, Mingze Xu, Hui Ding, Yuanjun Xiong, and Wei Xia.
\newblock Learning self-consistency for deepfake detection.
\newblock In {\em Proceedings of the IEEE/CVF international conference on
  computer vision}, pages 15023--15033, 2021.

\bibitem{twostream}
Peng Zhou, Xintong Han, Vlad~I Morariu, and Larry~S Davis.
\newblock Two-stream neural networks for tampered face detection.
\newblock In {\em IEEE conference on computer vision and pattern recognition
  workshops (CVPRW)}, pages 1831--1839. IEEE, 2017.

\bibitem{uia-vit}
Wanyi Zhuang, Qi Chu, Zhentao Tan, Qiankun Liu, Haojie Yuan, Changtao Miao,
  Zixiang Luo, and Nenghai Yu.
\newblock Uia-vit: Unsupervised inconsistency-aware method based on vision
  transformer for face forgery detection.
\newblock {\em arXiv preprint arXiv:2210.12752}, 2022.

\end{thebibliography}
}

\newpage
\setcounter{section}{0}
\renewcommand\thesection{\Alph{section}}
\section{Theoretical proof of the feature purity}
In this section, we present the theoretical analysis that higher feature purity (i.e., contains more task-relevant information) will help the generalization.

For the entire forgery detection task, we let $p(x, y)$ represent the ground-truth joint probability distribution corresponding to data $x$ and label $y$. $x \in \mathcal{X}$ and $y \in \mathcal{Y}$.
Ideally, we want to get a model $f(x ; \theta):\{\mathcal{X} ; \Theta\} \rightarrow \mathcal{Y}$, $\theta \in \Theta$, which minimizes the following objective function during the training process \cite{li2022learning}:
\begin{equation}\label{eq:f0}
\min _f F(f)=\int \mathcal{L}(f(x ; \theta), y) d p(x, y)
\end{equation}
where $\mathcal{L}$ is the loss function in the training.

However, in the actual training process, we cannot know the ground-truth probability distribution $p(x, y)$, but usually use a training set $D_{train}$ that we can obtain, and approximate Eq. (\ref{eq:f0}) through average calculation.
Let $I$ denote the number of data, the actual training target of the corresponding model $f_{1}(x ; \theta_1)$ is:
\begin{equation}\label{eq:fac}
\begin{gathered}
\min _{f_1} F_{actual}\left(f_1\right)=\frac{1}{I} \sum_{i=1}^I \mathcal{L}\left(f_1\left(x_i; \theta_1\right), y_i\right) \\
\text { s.t. }\left(x_i, y_i\right) \in D_{{train }}
\end{gathered}
\end{equation}


Comparing Eq. (\ref{eq:f0}) and Eq. (\ref{eq:fac}), the model $f_1$ obtained is not close to the ideal $f$ well due to the deviation of $D_{train}$ to $p(x, y)$ and the average approximation.
When $D_{train}$ and $p(x, y)$ are biased, model $f_1$ may satisfy the goal of Eq. (\ref{eq:fac}) by learning some ``shortcut features" \cite{shapes} which exist in the bias part and are not relevant to the forgery detection task.
Therefore, when faced with unseen domain data outside $D_{train}$, $f_1$ does not apply well, resulting in weak generalization. 
On the contrary, if we make the features of $f_1$ have as few forgery-irrelevant features as possible from the bias part (i.e., the feature purity is as high as possible), then $f_1$ will be more approximate to $f$, thus achieving better generalization.

\section{Solving details for Eq. (2)}
In this section, we show the details of solving Eq. (2) of the paper under the constraints of Eq. (1). 

As we mentioned in the paper, $\bm{g_r}$ represents the guide embedding of the real domain obtained by random initialization, and $\left\{\bm{g_{_{f_i}}}\right\}_{i=1}^N$ is the guide embedding of forgery domains that needs to be solved. 
We first use $\delta\left(\bm{g_{_{f_i}}}\right)$ to represent the constraints in Eq. (1) of the original paper:
\begin{equation}
\delta\left(\bm{g_{_{f_i}}}\right)=e^{\bm{g_r}^{T} \bm{g_{_{f_i}}}}-e^{\cos \left(\theta_0\right)}=0 \quad(i=1, \cdots, N)
\end{equation}
Then we aim to minimize Eq. (2) of the original paper, subject to the constraints of $\delta\left(\bm{g_{_{f_i}}}\right)$, which is formulated as:
\begin{equation}
\begin{gathered}
\min \ L\left(\left\{\bm{g_{_{f_i}}}\right\}_{i=1}^N\right)=\frac{1}{N} \sum_{i=1}^N \log \sum_{j=1}^N e^{\bm{g_{{}_{f_i}}}^T \bm{g_{_{f_j}}} / \tau} \\
\text { s.t. } \delta\left(\bm{g_{_{f_i}}}\right)=0 \quad(i=1, \cdots, N)
\end{gathered}
\end{equation}
We solve this based on the Lagrangian multiplier method \cite{lagrangian}. Let $\omega_i$ denote the Lagrangian multiplier, then the new solution function $\mathcal{H}(\cdot)$ can be constructed as:
\begin{equation}
\mathcal{H}\left(\left\{\bm{g_{_{f_i}}}\right\}_{i=1}^N\right)=L\left(\left\{\bm{g_{_{f_i}}}\right\}_{i=1}^N\right)+\sum_{i=1}^N \omega_i \cdot \delta\left(\bm{g_{_{f_i}}}\right)
\end{equation}
By calculating the partial derivatives of $\mathcal{H}$ to $\bm{g_{_{f_i}}}$ and $\omega_i$ and setting them to 0, $\left\{\bm{g_{_{f_i}}}\right\}_{i=1}^N$ can be obtained:
\begin{equation}
\nabla_{\bm{g_{_{f_i}}}} \mathcal{H}=\frac{\partial \mathcal{H}}{\partial \bm{g_{_{f_i}}}}=\nabla L+\omega_i \nabla \delta=\mathbf{0}
\end{equation}
\begin{equation}
\nabla_{\omega_i} \mathcal{H}=\frac{\partial \mathcal{H}}{\partial \omega_i}=\delta\left(\bm{g_{_{f_i}}}\right)=0
\end{equation}

\section{More details on hyper-parameters}
\noindent{\textbf{$\bm k$ in A-DBM:}}
{$k$ is $\left|K_i\right|$ in Eq. (5) of the paper. When $k$=10, 30, 50, 55, 60, 80, and 100, the AUCs on CelebDF are 79.35, 80.65, 83.02, 84.97, 84.91, 84.89, and 84.95. Stability is reached when $k$=55. When $k$=200, AUC drops to 81.96. In our training, each forgery domain has about 20,000 data, and the range of $55/20000$=0.275\% can be regarded as the nearest neighbor.
}

\noindent{\textbf{The number of clusters:}}
{In the decoupling module, we use clustering based on self-supervised features to explore potential similarities between data.
When the number of clusters is 100, 300, 500, 700, and 1000, the AUCs on CelebDF are 79.96, 81.78, 84.97, 83.65, and 83.42. When the number is small, it is easy to group less similar data into one cluster, and separating these data does not serve the purpose of decoupling irrelevant similarities well. When the number is large, it will cause similar data to be divided into different clusters, and when we conduct pushing operation, these data are not covered, so the performance will be reduced. When the number is 500, optimal performance is achieved.
}

\begin{table*}[t]
\centering 
\begin{tabular}{l|cccc|cccc}
\toprule[1.2pt]
Train   Set                      & \multicolumn{4}{c|}{DF   F2F NT}                                  & \multicolumn{4}{c}{DF F2F FS}     \\ \cmidrule(r){1-1} \cmidrule(r){2-5} \cmidrule(r){6-9}
Test Set                         & \multicolumn{2}{c}{FS (HQ)}     & \multicolumn{2}{c|}{FS (LQ)}    & \multicolumn{2}{c}{NT (HQ)}     & \multicolumn{2}{c}{NT (LQ)}     \\ \cmidrule(r){1-1} \cmidrule(r){2-3} \cmidrule(r){4-5} \cmidrule(r){6-7} \cmidrule(r){8-9}
                                 & Acc            & AUC            & Acc            & AUC            & Acc            & AUC            & Acc            & AUC            \\\cmidrule(r){2-3} \cmidrule(r){4-5} \cmidrule(r){6-7} \cmidrule(r){8-9}
$L_{ce-2}$                       & 76.61          & 85.89          & 91.82          & 96.99          & 79.02          & 87.28          & 81.93          & 90.28          \\
$L_{ce-(1+N)}$                   & 77.65          & 85.16          & 91.85          & 97.07          & 81.26          & 88.14          & 82.63          & 90.76          \\
$L_{guide}$                      & \textbf{78.44} & \textbf{86.95} & \textbf{92.33} & \textbf{97.34} & \textbf{82.07} & \textbf{89.13} & \textbf{83.95} & \textbf{92.04} \\ \bottomrule[1.2pt]
\end{tabular}
\vspace{0.2cm}
\caption{Comparisons of methods that increase the discrimination of different domains, including the results of FS and NT as the test set under the cross-test setting within FF++.}\label{tab:sup_loss}
\end{table*}

\begin{table*}[!ht]
\centering 
\begin{tabular}{l|cccc|cccc}
\toprule[1.2pt]
Train   Set                      & \multicolumn{4}{c|}{DF   F2F NT}                                  & \multicolumn{4}{c}{DF F2F FS}     \\ \cmidrule(r){1-1} \cmidrule(r){2-5} \cmidrule(r){6-9}
Test Set                         & \multicolumn{2}{c}{FS (HQ)}     & \multicolumn{2}{c|}{FS (LQ)}    & \multicolumn{2}{c}{NT (HQ)}     & \multicolumn{2}{c}{NT (LQ)}     \\ \cmidrule(r){1-1} \cmidrule(r){2-3} \cmidrule(r){4-5} \cmidrule(r){6-7} \cmidrule(r){8-9}
                                 & Acc            & AUC            & Acc            & AUC            & Acc            & AUC            & Acc            & AUC            \\\cmidrule(r){2-3} \cmidrule(r){4-5} \cmidrule(r){6-7} \cmidrule(r){8-9}
w/o $L_{guide}$        & 79.92          & 88.26          & 95.05          & 98.17          & 82.91          & 91.49          & 86.32          & 94.46        \\
w/o $L_{pull}$\&$L_{push}$ & 82.51          & 90.54          & 96.93          & 99.25          & 84.79          & 92.17          & 87.42          & 95.03        \\
w/o $L_{pull}$        & 83.48          & 92.69          & 97.05          & 99.31          & 86.94          & 93.86          & 88.56          & 95.91         \\
w/o $L_{push}$        & 84.35          & 93.07          & 97.24          & 99.40          & 87.30          & 94.83          & 88.67          & 96.14        \\
w/o {\small A-DBM} & 80.14          & 88.79          & 95.67          & 98.21          & 85.73          & 93.21          & 87.49          & 94.75        \\ \midrule[0.6pt]
ours           & \textbf{86.32} & \textbf{94.11} & \textbf{97.90} & \textbf{99.68} & \textbf{88.04} & \textbf{96.15} & \textbf{89.95} & \textbf{97.12} \\ \bottomrule[1.2pt]
\end{tabular}
\vspace{0.2cm}
\caption{Ablation performance after removing each module of the method, including the results of FS and NT as the test set under the cross-test setting within FF++.}\label{tab:sup_abla}
\end{table*}

\section{Computational cost}
For FLOPS, A-DBM calculates the nearest neighbor matrix, and this increase is 0.104\% of EN-B4 FLOPS, so the time consumption will not increase significantly. For memory consumption, the decoupling model needs to store a feature set $V$, and this increase is only 10M.
Our method focuses on the loss functions, so model parameters are not changed.

\section{Additional experiments}
In this section, we show more results of our method on cross-test setting within FF++ to demonstrate the effectiveness of our method in multiple experimental settings.
In the first two parts of this section, we show the ablation results of using FS and NT as the test set.
In the third part, we show the comparison with other recent methods.

\subsection{Methods to distinguish forgery domains}
For methods of enhancing the forgery domain discrimination, we regard results based on the binary cross-entropy loss $L_{ce-2}$ as the baseline. Based on this, we compare the multi-classification cross-entropy loss $L_{ce-(1+N)}$ that can also distinguish multiple forgery domains. The performance comparisons of $L_{ce-2}$, $L_{ce-(1+N)}$, and our $L_{guide}$ with FS and NT as the test set are shown in Table \ref{tab:sup_loss}.

Similar to the results on DF and F2F, on FS and NT, our $L_{guide}$ achieves the best performance among the three losses.
For example, on the NT dataset, the AUCs on HQ and LQ are 1.85\% and 1.76\% higher than $L_{ce-2}$, and 0.99\% and 1.28\% higher than $L_{ce-(1+N)}$, respectively.
For $L_{ce-(1+N)}$, it outperforms $L_{ce-2}$ in most cases, but on FS (HQ), its AUC is 0.73\% lower than $L_{ce-2}$. This shows that it is not feasible to simply regard distinguishing different domains as an ordinary multi-classification task.
To improve generalization, we need to keep the real domain far enough away from forgery domains to cope with the complexity of the forgery domain, while also ensuring the distinction between the forgery domains. That is, the separation degree between real and forgery should be much larger than the degree between forgery and forgery. Our guide-space based method does this well and thus achieves good performance.

\begin{figure*}[t]
\centering\includegraphics[width=0.96\textwidth]{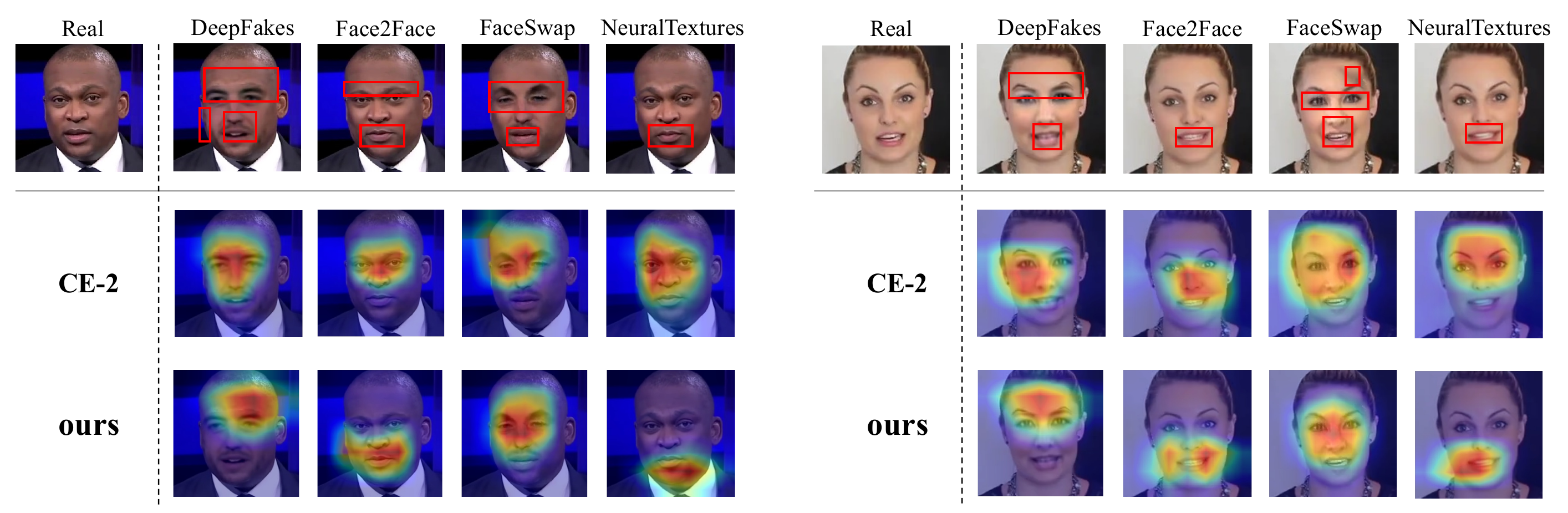}
\caption{The heatmap comparisons of binary cross-entropy (CE-2) and our method. Forgery artifacts are marked in red frames.}
\label{fig:sup_heat}
\end{figure*}

\subsection{Importance of different modules}
Table \ref{tab:sup_abla} lists the performance of our method on FS and NT as the test set when each key module of our method is removed respectively.
It can be seen that each module contributes to the overall performance, and its removal will lead to a decrease in performance.
Both $L_{guide}$ and $L_{pull}$\&$L_{push}$ can achieve the separation of different domains and the aggregation of the same domain, but removing $L_{guide}$ has a greater impact. This is because guide embeddings can achieve the controllability of separation and aggregation, and the decoupling model enhances this discriminativeness by reduce the interference of irrelevant similarities between domains.
For the A-DBM module, it has different influences on different datasets. For example, on FS (HQ), removing it will reduce AUC by 5.32\%, and on NT (LQ), AUC will decrease by 2.37\%. Overall, A-DBM focuses on weak samples in the optimization process and plays an important role in the overall performance.

\subsection{Cross test on FF++}
In cross-test setting within FF++, we compare the performance of our method and the recent methods. In Table \ref{tab:sup_cross}, we compare the results of DCL \cite{dcl}, Face X-ray \cite{facexray}, and Xception\cite{xception}.
It can be seen that under DF, F2F, FS, and NT, our method achieves optimal performance. Under NT, DCL \cite{dcl} achieves the sub-optimal performance, and ours is 2.3\% higher than it.

\begin{table}[t]
\centering 
\begin{tabular}{p{2.4cm}|p{1.3cm}<{\centering} p{1.3cm}<{\centering} p{1.2cm}<{\centering} p{0.8cm}<{\centering}}
\toprule[1.2pt]
Training Set   & \multicolumn{4}{c}{Train on remaining three}                                                                                 \\ \cmidrule(r){1-1} \cmidrule(r){2-5}
Testing Set    & \multicolumn{1}{c|}{DF}            & \multicolumn{1}{c|}{F2F}           & \multicolumn{1}{c|}{FS}            & NT            \\ \hline
Xception \cite{xception}       & \multicolumn{1}{c|}{93.9}          & \multicolumn{1}{c|}{86.8}          & \multicolumn{1}{c|}{51.2}          & 79.7          \\
Face X-ray \cite{facexray}     & \multicolumn{1}{c|}{99.5}          & \multicolumn{1}{c|}{94.5}          & \multicolumn{1}{c|}{93.2}          & 92.5          \\
DCL \cite{dcl} & \multicolumn{1}{c|}{95.7}          & \multicolumn{1}{c|}{98.2}          & \multicolumn{1}{c|}{91.5}          & 93.9 \\ \hline
Ours           & \multicolumn{1}{c|}{\textbf{99.8}} & \multicolumn{1}{c|}{\textbf{98.9}} & \multicolumn{1}{c|}{\textbf{94.1}} & \textbf{96.2}        \\ \bottomrule[1.2pt]
\end{tabular}
\vspace{0.2cm}
\caption{Cross-test within FF++ (HQ). Generalization performance AUC (\%) when testing on one type after training on the remaining three types.}\label{tab:sup_cross}
\end{table}

\section{More visualizations}
In this section, we show more heatmaps of binary cross-entropy (CE-2) and our method, and these visualizations are shown in Figure \ref{fig:sup_heat}.

Similar to the results shown in Figure 6 of the paper, for CE-2, there are certain similarities in the areas that the models focus on under different forgery types, and they are concentrated in the central area of the face. While the areas that our method focuses on are the respective artifacts corresponding to different forgery types.
For face-swapping methods (DeepFakes and FaceSwap) that replace the whole face, it is reasonable for the model to focus on either the central area of the face or the boundary artifacts.
For the face reenactment methods (Face2Face and NeuralTextures), the forgery traces are mainly in local areas such as the mouth and eyes. But due to the interference of forgery-irrelevant similarities between different forgery methods, CE-2 still focus on the central area of the face similar to the face-swapping methods, and does not extract the distinguishable features of F2F and NT well. In contrast, our method can pay attention to the corresponding forgery traces and extract better forgery-related features.
\end{document}